%% file: main.tex
\pdfoutput=1

\documentclass[11pt]{article}

\usepackage{acl}

\usepackage{times}
\usepackage{latexsym}

\usepackage[T1]{fontenc}

\usepackage[utf8]{inputenc}

\usepackage{microtype}

\newcommand{\papertitle}{
  Jointly Reparametrized Multi-Layer Adaptation for\\Efficient and Private Tuning
}
\title{\papertitle}

\author{Umang Gupta\\
USC Information Sciences Institute\\
{\texttt{umanggup@usc.edu}}
\And Aram Galstyan \\
USC Information Sciences Institute\\
\AND Greg {Ver Steeg} \\
University of California Riverside \\
}

\usepackage{booktabs}
\usepackage{diagbox}
\usepackage{graphicx}
\usepackage{amsmath}
\usepackage{amsthm}
\usepackage{amsfonts}
\usepackage[]{xcolor}
\usepackage{colortbl}

\usepackage[utf8]{inputenc} %
\usepackage{hyperref}       %
\usepackage{url}            %
\usepackage{cleveref}
\usepackage{nicefrac}       %
\usepackage{microtype}      %
\usepackage{makecell}
\usepackage{multirow}
\usepackage{subcaption}
\usepackage{pgfplots}
\pgfplotsset{compat=1.18}

\usepackage{todonotes}
\usepackage{pifont}%
\newcommand{\cmark}{\ding{51}}%
\newcommand{\xmark}{\ding{55}}

\newcommand{\bert}{\texttt{BERT}}
\newcommand{\roberta}{\texttt{RoBERTa}}
\newcommand{\bertb}{\texttt{BERT-base}}
\newcommand{\robertab}{\texttt{RoBERTa-base}}

\newcommand{\robertal}{\texttt{RoBERTa-large}}
\newcommand{\cls}{\texttt{[CLS]}}
\newcommand{\mask}{\texttt{[MASK]}}
\newcommand{\warp}{{WARP}}
\newcommand{\adapters}{Adapter}
\newcommand{\lora}{LoRA}
\newcommand{\bitfit}{BitFit}
\newcommand{\compacters}{Compacter}
\newcommand{\layershift}{SLaSh}
\newcommand{\latentprompt}{JR-WARP}
\newcommand{\result}[2]{#1\textsubscript{$\pm$#2}}
\newcommand{\zvector}{\ensuremath{\mathbf{z}}}
\newcommand{\randomW}{\ensuremath{\mathbf{W}}}
\newcommand{\layervariable}{\ensuremath{l}}
\newcommand{\zdim}{\ensuremath{d}}
\newcommand{\biasdim}{\ensuremath{d'}}
\newcommand{\numclass}{\ensuremath{C}}
\newcommand{\numlayers}{\ensuremath{L}}
\definecolor{ForestGreen}{RGB}{34,139,34}
\definecolor{lightgray}{rgb}{0.84, 0.84, 0.84}
\definecolor{cadmiumgreen}{rgb}{0.0, 0.42, 0.24}
\newcommand{\tblrowcolor}{lightgray}
\newcommand{\tblrowcolorname}{gray}

\newcommand{\seperator}{}

\newcommand{\ie}{\emph{i.e.}}
\newcommand{\eg}{\emph{e.g.}}
\newcommand{\vs}{vs.\ }
\newcommand{\real}{\ensuremath{\mathbb{R}}}
\newcommand{\normal}{\ensuremath{\mathcal N }}
\newcommand{\uniform}{\ensuremath{\mathcal U }}
\newcommand{\highlight}[1]{\underline{\textit{#1}}}

\begin{document}
\maketitle

\input{sections/abstract}

\input{sections/introduction.tex}

\input{sections/approach.tex}

\input{sections/results.tex}

\input{sections/related.tex}

\input{sections/conclusion.tex}

\input{sections/limits.tex}

\input{sections/ethics.tex}

\bibliography{anthology,custom}
\bibliographystyle{acl_natbib}

\input{appendix/main}

\end{document}

%% file: sections/abstract.tex
\begin{abstract}

  Efficient finetuning of pretrained language transformers is becoming increasingly prevalent for solving natural language processing tasks. While effective, it can still require a large number of tunable parameters. This can be a drawback for low-resource applications and training with differential-privacy constraints, where excessive noise may be introduced during finetuning. To this end, we propose a novel language transformer finetuning strategy that introduces task-specific parameters in multiple transformer layers. These parameters are derived from fixed random projections of a single trainable vector, enabling finetuning with significantly fewer parameters while maintaining performance. We achieve within 5\% of full finetuning performance on GLUE tasks with as few as 4,100 parameters per task, outperforming other parameter-efficient finetuning approaches that use a similar number of per-task parameters. Besides, the random projections can be precomputed at inference, avoiding additional computational latency. All these make our method particularly appealing for low-resource applications. Finally, our method achieves the best or comparable utility compared to several recent finetuning methods when training with the same privacy constraints, underscoring its effectiveness and potential real-world impact.

\end{abstract}

%% file: sections/introduction.tex
\section{Introduction}\label{sec:intro}
\input{figures/intro_fig}

Transformer-based bidirectional language models (LMs), pretrained on a sizeable text corpus and finetuned on task-specific objectives, outperform models trained from scratch by large margins~\cite{devlin-etal-2019-bert,liu2019roberta}.
The straightforward approach to finetune a language model is to initialize with pretrained parameters and train the model on the downstream task. However, it is inefficient to finetune language models for each task as it requires training and storing a massive number of parameters per task (roughly the same as the size of language models)~\cite{radford2019language,devlin-etal-2019-bert}. These inefficiencies are exacerbated in resource-constrained settings, such as personal devices with limited or federated learning scenarios where the costs of communicating parameter updates
 may
 limit the scope of applications~\cite{xu2022training,ro2022scaling}.

The shortcomings of naive finetuning methods have motivated research into approaches that identify and train fewer task-specific parameters~\cite{treviso2022efficient}. Those parameter-efficient finetuning methods work by introducing task-specific trainable layers while freezing most of the pretrained language model parameters (\eg, \adapters~\cite{houlsby2019parameter,pfeiffer-etal-2021-adapterfusion}, \lora~\cite{hu2022lora}) or by introducing task-specific trainable prompts or inputs (\eg, prompt-tuning based \warp~\cite{hambardzumyan-etal-2021-warp}, prefix-tuning~\cite{li-liang-2021-prefix}). We summarize the key properties of prominent efficient finetuning methods in \Cref{tab:properties}.
Among these methods,  \warp\ is particularly interesting. It demonstrated comparable performance to full-finetuning with as few as 25K trainable parameters on natural language understanding (NLU) tasks.

\input{tables/properties.tex}

\warp\ inserts trainable token embeddings around input, \ie,  task-specific parameters are inserted only in the input layer. Due to this,  \warp\ is limited compared to other methods that insert trainable parameters in different layers (\ie, \highlight{Multi-layer}), as the information may not propagate correctly to the deeper layers~\cite{liu-etal-2022-p}.
As such, our proposed method
inserts task-specific information in each transformer block. In particular, we add a bias or shift vector to the output feed-forward layer's activation in each transformer block. All these shifts are derived from a single trainable vector, keeping the total trainable  parameter count similar to \warp.

This is in contrast to \bitfit~\cite{ben-zaken-etal-2022-bitfit}, which updates all the bias parameters independently without sharing. Our proposed \highlight{parameter sharing} or joint reparametrization of task parameters drastically reduces the number of trainable parameters without significant performance degradation. On average,
our method is within two points of \bitfit\ on NLU tasks but uses 20x fewer parameters. Specifically,  we achieve within 5\% of full finetuning performance with only 4.1K parameters (see~\Cref{fig:intro}), outperforming \warp\ which uses a similar number of parameters. Lastly, we show that parameter sharing and multi-layer tuning  can also improve
\warp.

\warp\ increases the effective
sequence length, and
\adapters\ inserts task-specific layers,  incurring additional computational overhead. In contrast, our method is \highlight{efficient} in memory usage and run-time during training. Further, task-specific parameters learned by our approach can be fused with LM during inference, leading to no additional  latency during inference, making it especially appealing for resource-constrained applications. Besides computational efficiency, our approach's  parameter efficiency  makes it an excellent private learner.
Our approach's utility is competitive or outperforms the best differential private finetuning results~\cite{yu2022differentially}
when training for similar levels of privacy.

%% file: figures/intro_fig.tex
\begin{figure}
    \centering
        \includegraphics[width=0.4\textwidth, trim={0cm 0 0 0\textheight},clip]{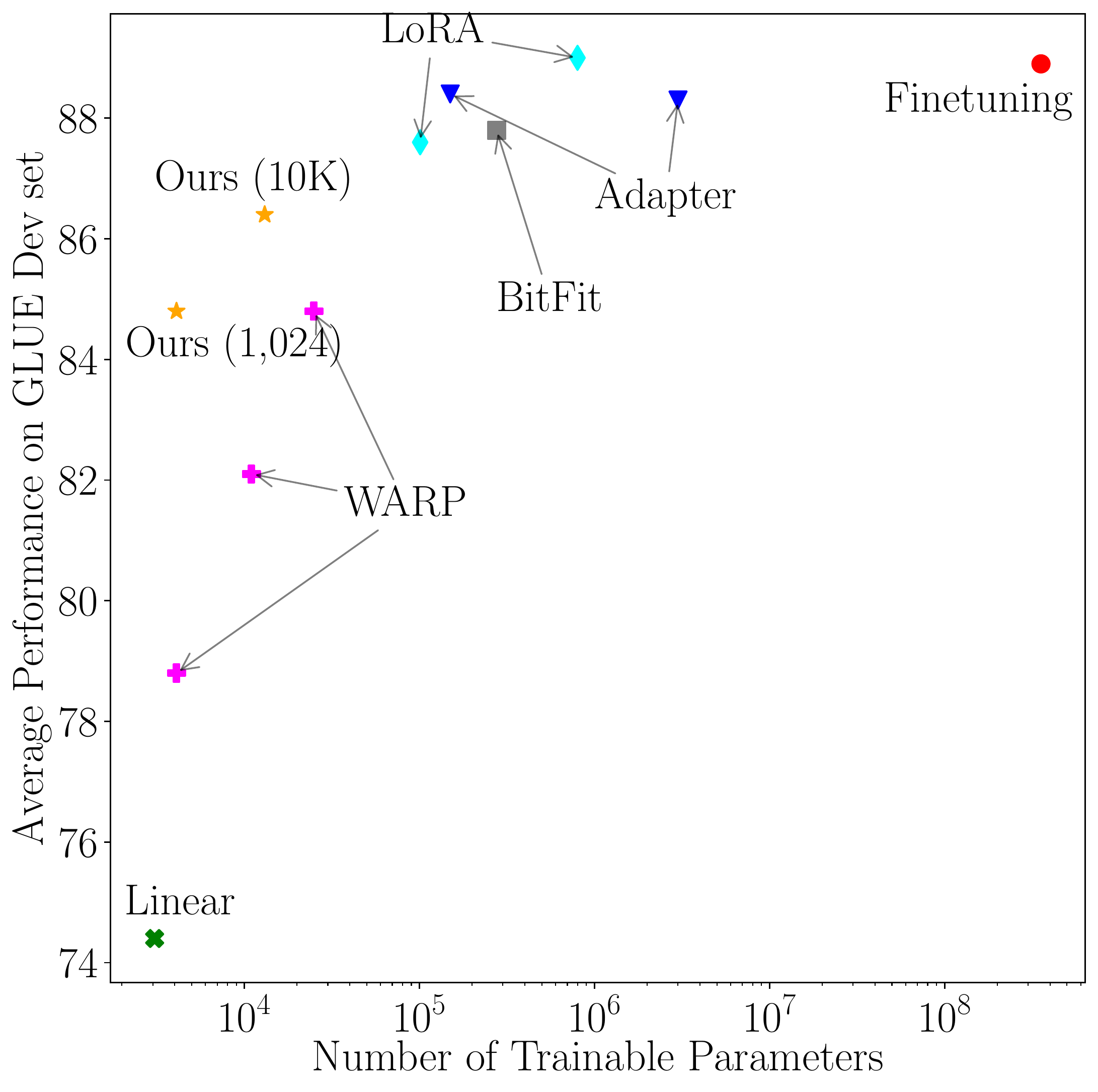}
    \caption{Performance \vs Parameters trade-off
    on GLUE benchmark with pretrained \robertal.}
    \label{fig:intro}
\end{figure}

%% file: tables/properties.tex
\newcommand{\tick}{{\color{cadmiumgreen}\cmark}}
\newcommand{\cross}{{\color{red}\xmark}}

\begin{table}
    \centering
    \fontsize{10}{11}\selectfont
    \setlength{\tabcolsep}{4pt}
    \begin{tabular}{l c c c}
        \toprule
        \multirow{2}{*}{Method}
            & \makecell[tc]{Parameter \\Sharing}
            & \makecell[tc]{Efficient\\Inference}
            & \multirow{2}{*}{Multi-layer} \\
            \cmidrule(r){1-1}
            \cmidrule(lr){2-2}
            \cmidrule(lr){3-3}
            \cmidrule(l){4-4}
        Adapter        & \cross & \cross  & \tick  \\
        LoRA           & \cross & \tick   & \tick  \\
        BitFit         & \cross & \tick   & \tick  \\
        WARP           & \cross & \cross  & \cross \\
        Ours           & \tick  & \tick   & \tick  \\
        \bottomrule
    \end{tabular}
    \caption{Parameter Efficient Finetuning Methods}
    \label{tab:properties}
\end{table}

%% file: sections/approach.tex
\seperator

\section{Method}\label{sec:approach}
\input{figures/method_fig.tex}

\paragraph{Model.}
\Cref{fig:approach} summarizes our model, highlighting task-specific parameters with colored fonts.
Specifically, we consider a trainable vector
${\color{red}\zvector}\in \real^\zdim$
to incorporate task-specific information in each transformer block.
We do so by projecting {\color{red}\zvector} with random but fixed matrices ${\color{ForestGreen}\randomW_{\layervariable}}$ to obtain shift vectors ${\color{blue}\zvector_\layervariable}$ for the $\layervariable$\textsuperscript{th} transformer block
(${\color{blue}\zvector_\layervariable}\in \real^{\biasdim_\layervariable}$,  ${\color{ForestGreen}\randomW_{\layervariable}} \in \real^{\biasdim_\layervariable\times\zdim}$, and $l\in \{1\ldots \numlayers\}$).  ${\color{blue}\zvector_\layervariable}$\
is added to the output activations of the respective transformer block, as shown in \Cref{fig:approach}. ${\color{blue}\zvector_\layervariable}$
is of the same dimensionality as the activations
of the output feed-forward layer in the
$\layervariable$\textsuperscript{th} transformer block ($\biasdim_\layervariable$), and ${\color{red}\zvector}$\ is shared between all the blocks. Hence, we call our approach Shared Layer Shift or  \highlight{\layershift}.

The random projection matrices, ${\color{ForestGreen}\randomW_{\layervariable}}$, are not trainable and are fixed throughout the training. We initialize ${\color{ForestGreen}\randomW_{\layervariable}}$\  and {\color{red}\zvector}\  with zero-centered Gaussian or Uniform distribution for our experiments (See \Cref{sec:init_ablations} for ablations on initialization choices).

\layershift\ is akin to training only bias parameters of the output feed-forward layers. However, the projection step decouples the dimensions of {\color{red}\zvector} and activations, providing the flexibility to change the number of trainable parameters and control the complexity of the model by varying \zdim\ irrespective of the activation dimensions.
Our choice of adding ${\color{blue} \zvector_\layervariable}$ to only output activations is inspired by \citet{subramani2020discovering}, who use a similar setup to learn sentence representations. We also consider adding the shifts to other activations, such as intermediate activations or activations after the self-attention layer in \Cref{sec:pos_ablations}. In particular, adding shifts to output activations performs similarly or better than other choices. Adding shifts to intermediate layers performs similarly to adding shifts to the output layer. However, the dimensionality of intermediate activations is usually more than that of output activations which would increase the size of projection matrices, making it an inferior choice.

\paragraph{Classification Head.} We experiment with token classification and sequence classification tasks with \bert-like models. To this end, we remove the decoder layer of the pretrained LM and attach a task-specific
linear layer ({\color{red} Classifier}) to predict the output from text representations. Verbalizers~\cite{schick-schutze-2021-just} can also be used.

\paragraph{Number of Parameters.}
\layershift\ only trains the task-specific vector (\zvector) and the prediction head (Classifier), usually a classification or regression layer. Suppose the number of class labels is \numclass.  \layershift\ will only use
$\zdim+ \numclass\times(\biasdim_\numlayers+1)$
trainable parameters per task, where
$\biasdim_\numlayers$
is the activation dimension of the last transformer block. 
In our implementation, we maintain additional
$\sum_{\layervariable=1}^\numlayers \biasdim_\layervariable \times\zdim$
parameters for $\randomW_l$ matrices during training. However,  these matrices can also be generated on the fly from the random seed or  state of the random number generator for both backward and forward pass computation.
More concretely, \robertal\ has $\numlayers=24, \biasdim_l=1024 \ \ \forall l \in \{1\ldots \numlayers\}$, and for GLUE tasks, the number of classes, $\numclass$,  can be 3 maximum. If \zdim\ is set to 1,024, only 4,099 trainable parameters are required per task. In contrast, \robertal\ has  {355M} parameters.

The maximum size of \zvector\ could be the sum of the dimensions of all the shift vectors, \ie, $\sum_{\layervariable=1}^L \biasdim_\layervariable$. Increasing the size beyond that is similar to training respective bias parameters independently without any sharing or reparametrization.

\paragraph*{Inference.}
Pretrained LM parameters are shared across all the tasks. The projection weights remain unchanged during the
training and can be reproduced from the random seed or random number generator's state. Hence, once the model is trained, only  \zvector\ and classifier parameters   need to be preserved.
Our approach maintains computational efficiency during inference as it does not require additional computations apart from the language model inference. Indeed, once the shift vectors $\zvector_l$ are computed, they can be combined with biases of the output feed-forward layers.%

\paragraph{Improving Prompt-Tuning.} These joint reparametrization of task parameters can also improve prompt-tuning methods such as \warp. We make two modifications --- a) Insert prompts in different layers, and b) Prompts are derived from a single vector. We refer to this as \latentprompt\ (Jointly Reparametrized WARP. We provide more details about \latentprompt\ in \Cref{sec:latent_prompt}.  Multilayer or deep-prompts have already been shown improve performance~\cite{liu-etal-2022-p, li-liang-2021-prefix}. Here we improve  parameter efficiency while maintaining performance.

%% file: figures/method_fig.tex
\begin{figure*}
    \centering
    {
        \includegraphics[width=0.85\textwidth,clip, trim={10 11.9 30 4}]{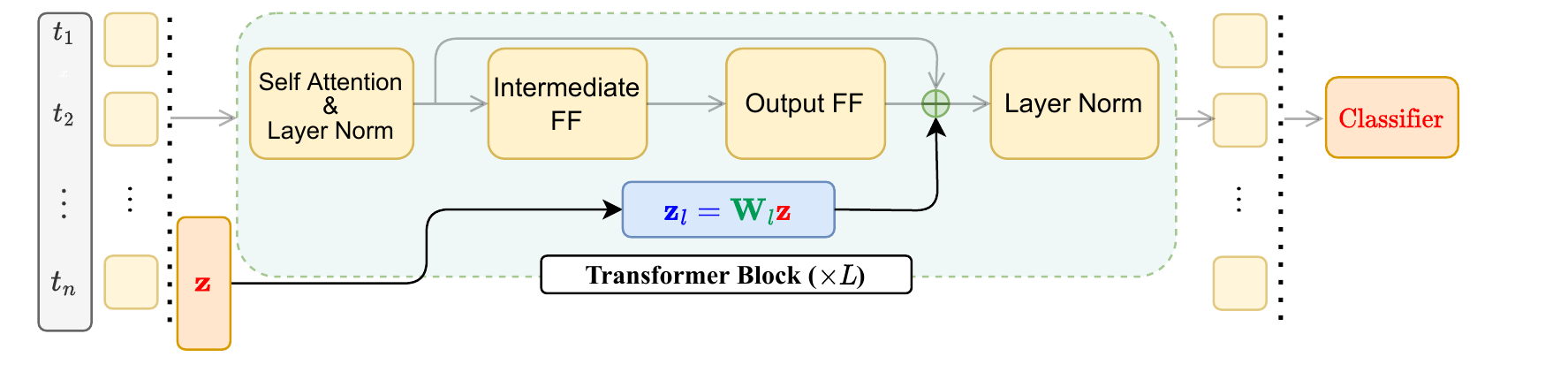}
    }
    \caption{Shared Layer Shift or \layershift\ for sequence classification tasks.  The shifts are obtained by the  projections
    \protect{${\color{ForestGreen}\randomW_\layervariable}{\color{red}\zvector}$}
    and added to the corresponding transformer block's output activation. ${\color{red}\zvector}$ is shared across all the blocks. Red font indicates trainable modules, \ie, parameters that are updated during finetuning. Other parameters remain unchanged during the finetuning. $[t_i]_{i=1}^n$ denotes the sequence of tokens. }
    \label{fig:approach}
\end{figure*}

%% file: sections/results.tex
\seperator

\section{Experiments}\label{sec:experiments}

We evaluate our  approach for sequence classification tasks in \Cref{sec:seq_clf} with the General Language Understanding Evaluation (GLUE) benchmark~\cite{wang2018glue} and token classification tasks with named entity recognition (NER) on CoNLL 2003 dataset~\cite{tjong-kim-sang-de-meulder-2003-introduction} in \Cref{sec:token_clf}. We report memory and training time requirements to quantify the computational efficiency
in \Cref{sec:profiling}. Finally, we demonstrate the utility of our approach for differential private finetuning of LMs in \Cref{sec:dp}.\footnote{Details about training, hyperparameter search, and best hyperparameters for all the experiments are in \Cref{sec:hyperparams}. The code is available at \url{https://github.com/umgupta/jointly-reparametrized-finetuning}.}

\input{tables/glue_results.tex}

\paragraph*{Baselines.}
We compare  against full-finetuning and several prominent parameter-efficient finetuning techniques. Specifically, we compare with \adapters~\cite{houlsby2019parameter}, Low-Rank Adaptation (\lora, \citet{hu2022lora}), \bitfit~\cite{ben-zaken-etal-2022-bitfit}, and Word Adversarial Reprogramming (WARP, \citet{hambardzumyan-etal-2021-warp}).

\highlight{\adapters}\ introduces task-specific feed-forward layers in each transformer block. \adapters\ typically trains down-project and up-project feed-forward layers in pairs for each transformer block. The dimensions of the down-projection (denoted as $m$) govern the per-task trainable parameters.

\highlight{Low-rank adaptation}, or \highlight{\lora} learns the change in the pretrained weights, \ie, $\Delta W$, for the downstream tasks. $\Delta W$ is parameterized as the product of low-rank matrices, which requires much fewer parameters than full-finetuning. The rank of the matrices determines per-task parameters.

\highlight{\warp\textsubscript{n}} introduces $n$ learnable input tokens  by adding trainable embeddings to the input. It is the continuous version of prompt-tuning and a special case of PrefixTuning~\cite{li-liang-2021-prefix}, with prefixes  introduced only in the embedding layer. The learned tokens  do not necessarily correspond to an existing token from the vocabulary.

Finally, we compare with \highlight{\bitfit}, which finetunes only  all the bias parameters. Indeed, \bitfit\ finetunes a superset of parameters considered by our approach.
Further, \layershift\ shares trainable parameters across all the blocks, which is more efficient.

\input{tables/glue_results_footnote.tex}

\subsection{Sequence Classification Tasks}
\label{sec:seq_clf}

\input{tables/glue_roberta_base.tex}

\paragraph{Datasets.}
We use the GLUE benchmark for sequence classification. We consider 2 single-sentence tasks  and 6 sentence pair tasks from the GLUE benchmark. Corpus of Linguistic Acceptability (CoLA) and Stanford Sentiment Treebank (SST-2) are the single sentence tasks, and the task is to predict grammatical acceptability and sentiment.
Microsoft Research Paraphrase Corpus (MRPC), Semantic Textual Similarity Benchmark (STS-B), and Quora Question Pairs (QQP) are the sentence similarity tasks. Multi-genre Natural Language Inference (MNLI), Question-Answering NLI (QNLI), and Recognizing textual entailments (RTE) are textual entailment prediction tasks. Similar to \citet{devlin-etal-2019-bert,houlsby2019parameter}, we omit results on Winograd Schema Challenge (WNLI)  as LMs do not outperform random prediction baselines.

All the tasks except STS-B are considered supervised classification tasks. Labels for STS-B are similarity scores from 1-5, and thus it is considered a regression task.
We report accuracy on matched validation set for MNLI, Matthew's correlation and Pearson correlation on CoLA and STS-B, F1-score for MRPC and QQP, and accuracy for the rest of the tasks on the development set. %
Model selection is also performed based on these metrics.

\paragraph{\cls\ \vs \mask\ Representations.}
We consider two sentence-level representations for sequence classification tasks --- \cls\ and \mask\ token representations. Masked language models (MLMs) such as \bert\ and \roberta\ are pretrained by attaching a \cls\ token to the beginning of the input text. The \cls\ token representation is trained with the next sentence prediction loss and thus touted as the sentence-level representation. To this end, most previous works use \cls\ token representations.
However, \citet{hambardzumyan-etal-2021-warp} suggested that \mask\ tokens representations, \ie, inserting the \mask\ token at the end of input for single-sentence or between the sentences for tasks involving sentence pairs, produces better results than using \cls\ token representation.

We also find that the \mask\ representations are better than \cls\ representations generally and report results with \mask\ representations in the paper. We compare the two in \Cref{sec:mask_vs_cls}.

\paragraph*{Training.} We use \roberta~\cite{liu2019roberta} as the pretrained model to  compare with previous works. For \layershift, we vary the number of parameters by varying the size of the \zvector\ vector. The output activation and embedding dimensions  are 1,024  in \robertal.  So, we train with \zdim\  = 1,024 and 2,048 to compare head-to-head with \warp.
We report results with \zdim\  = 5K and 10K for \robertab\ and \robertal, which improves the results further. To demonstrate the capabilities of  tuning only output activation's biases, we train with the maximum possible \zdim, \ie, the total number of activations, 9,216 and 24,576 for \robertab\ and \robertal. We also train \lora\ and \adapters\ with minimum parameter configurations (rank = 1 and m = 1) as the  results reported in their papers use a larger number of parameters than those concerning this work. We demonstrate that parameter sharing can also improve \warp\ by introducing \latentprompt\ and training it with \zdim\  = 5K and 10K for respective \roberta\ models.

\paragraph{Results.}  \Cref{tab:roberta_large_glue_result,tab:roberta_base_glue_result} summarize the results of finetuning with different methods using pretrained \roberta\ models.  Parameter-efficient finetuning approaches aim to achieve performance at par with full finetuning while using fewer parameters.  To this end, \Cref{fig:intro} provides a visual summary of the parameter \vs performance trade-offs.

\layershift's average performance is already within 4 points of full finetuning for both \robertab\ and \texttt{-large} models with \zdim\ = 1,024. This gap is further reduced by increasing the dimension of the \zvector\ vector. Even though the best models from our approach do not match the full-finetuning performance overall,  for smaller datasets such as STS-B, MRPC, and RTE, \layershift\ is competitive with full-finetuning. In the case of \robertal, we have 92.4 \vs 91.5 for STS-B, 90.9 \vs 91.3 for MRPC, and 86.6 \vs 84.1 for RTE with finetuning and \layershift, respectively.\footnote{Note that we consider the average performance of \layershift\ across different training runs, whereas, for baselines, performance from a single training run with fixed seed is reported. This can slightly exaggerate baseline numbers.} The parameter sharing reduces the per-task parameters considerably (4 orders of magnitude less) and is faster and more
efficient to train (\Cref{sec:profiling}). All these make our approach  suitable for low-resource, low-data applications such as training on edge devices or learning personalized models.

Most efficient tuning techniques tune a few hundred thousand parameters, except for \warp. It adds trainable parameters around input embeddings, which facilitates training with a few thousand parameters and is most comparable to our approach  in per-task parameters.
Our approach with \zdim\ = 2,048 (\ie, 5.1K parameters) outperforms \warp\ with 25K parameters on all datasets with less than 10K training samples. Further, \layershift\ outperforms the best results of \warp\ while using less than 60\% of parameters  (13K \vs 25K). These observations do not change even with \warp\   pretraining on the MNLI task to improve the performance on smaller datasets (\warp\textsubscript{MNLI}). We do not require this supervised pretraining trick.  These results validate the intuition that instead of introducing task parameters closer to the input layer as in \warp, it may be more effective to introduce the parameters throughout the layers as in \layershift.

Armed with this intuition, we improve \warp's performance  by introducing prompts in all transformer blocks derived from a single vector (\latentprompt). On average, it underperforms \layershift, and the variance among different training runs is higher. Nevertheless, \latentprompt\ performs comparably to \warp\textsubscript{20} (84.4 \vs 84.8) while using fewer parameters (13K \vs 25K), suggesting that reusing parameters across layers improves parameter efficiency but does not deteriorate  performance.

Next, we compare with \lora\ and \adapters, arguably the most prominent language transformer finetuning approaches.   We note that the \adapters\ (rank = 1) has a slightly better average performance than \lora\ (m = 1) (\Cref{tab:roberta_large_glue_result,tab:roberta_base_glue_result}).
\layershift\ performs comparably to these methods for smaller datasets,  using 5x fewer parameters and being roughly 2x faster to train for \robertab\ and 7x fewer parameters and roughly 1.25x faster to train for \robertal\ (\Cref{tab:roberta_large_glue_result,tab:roberta_base_glue_result,tab:profiling}). {For example, in the case of \robertab, we have 91.0 \vs 89.9 for STS-B, 92.3 \vs 90.7 for MRPC, and 78.3 \vs 76.7 for RTE with \adapters\ and \layershift, respectively.}

Finally, \layershift\ performs comparably to \bitfit\ while tuning much fewer parameters. As with the other baselines, it is only for the larger datasets that \bitfit\  considerably outperforms  \layershift. Further, we observe that tuning only output activation's biases, which used fewer than  15\% of \bitfit's parameters, performs comparably to \bitfit\ on average (last row of \Cref{tab:roberta_large_glue_result,tab:roberta_base_glue_result}).

Another interesting result is the performance of \bitfit\ \vs \adapters\ and \lora\ with a similar number of trainable parameters. We observe that \adapters\ and \lora\  outperform \bitfit\ on most tasks with fewer trainable parameters. For instance, \bitfit\ outperforms \lora\ on QNLI, CoLA, STS-B, MRPC with \robertal, and only STS-B and MRPC with \robertab. \adapters\ outperforms \bitfit\ on all the tasks with both pretrained models except MRPC with \robertal. These results contradict \citet{ben-zaken-etal-2022-bitfit}, suggesting that while tuning bias parameters may achieve close to finetuning, \lora\ or \adapters\ may yield better performance with fewer parameters.

\subsection{Token Classification Task}
\label{sec:token_clf}

Next, we evaluate our method on more complex token classification tasks such as NER. We consider the CoNLL-2003 (English) dataset. We use \bertb\texttt{-cased}\ as the pretrained-LM and finetune it to predict the 9 entity classes. We use the validation set for model selection and report micro-F1 on the test and validation sets.

\input{tables/ner_results.tex}

\input{tables/profiling}

\paragraph*{Results.}
Table~\ref{tab:ner_result} reports the results of finetuning with \bertb\texttt{-cased}\ for the NER task. We see similar trends in performance as the sequence classification task. However, owing to the complexity of the NER task, all the methods underperform full-finetuning significantly (91.35 F1 score). \layershift\ with 8K parameters underperforms full-finetuning by more than 4 points (86.49). The performance is improved to 88.30 by increasing the number of trainable parameters. However, \lora, \adapters, and \bitfit\ outperform the best results from \layershift\ by roughly 1.5 points but  use more than 3.5x parameters compared to \layershift. Among the parameter-efficient techniques, \adapters\ performed the best while using fewer parameters than \bitfit. Similar to \Cref{sec:seq_clf}, \layershift\ and \latentprompt\ outperform \warp. Hyperparameter tuning (\eg, increasing the sequence length)  can improve \latentprompt\ results further. Overall, \layershift\ is suitable for extremely low-parameter applications, even the token classification tasks, but it may degrade
 performance.

\subsection{Time \& Memory Requirements}
\label{sec:profiling}
One of the goals of parameter-efficient tuning is to achieve as much utility as possible while being efficient with memory and computing. To this end, we report memory and time for training 1 epoch on the QNLI dataset in \Cref{tab:profiling}.
Full finetuning requires longer execution time and more memory than any other approach, making a clear case for parameter-efficient approaches.  \layershift\ requires considerably less time and memory than \lora\ and \adapters\ --- 40\% less time and 33\% less memory for \robertab\, and 12\% less time and 30\% less memory for \robertal. The gains are less pronounced for \texttt{large} models than \texttt{base} because relatively more resources are utilized for transformer computations than tuning-specific computations.  Compared to \bitfit, \layershift\ trains faster, but the memory requirements are similar due to \layershift\ maintaining projection matrices during training.

{We maintained projection matrices in memory instead of generating them on the fly for our experiments, and \Cref{tab:profiling} uses this implementation. However, matrices can be generated on the fly for both forward and backward passes from the state of the random number generator, leading to a further reduction in memory usage. With this improvement, the memory usage comes down to 8.3 GB and 3.1 GB for the large and base model without significantly impacting training time.} 
Finally, \warp's memory utilization is identical to \layershift, but has slightly higher training time due to increased sequence length. \layershift\ is much more resource-efficient during training than other methods without too much compromise on performance.

Inference times for all the methods were similar. The time to perform inference over the QNLI validation set (5,463 examples) varied between 13.9-14.5 seconds for \robertab\ and 39.7-40.8 seconds for \robertal.

\input{tables/glue_results_pvt.tex}

\subsection{Differential Private Finetuning}
\label{sec:dp}
As machine learning is beginning to be applied in commercial settings and on user data, ensuring the privacy of training data is becoming crucial. Neural networks trained without safeguards can easily leak information about their private training data~\cite{carlini2021extracting,carlini2022quantifying}. To mitigate these issues, neural networks can be trained with a strong notion of privacy, Differential Privacy (DP), which limits the influence of a single training example on the result~\cite{dwork2014algorithmic}.

Differential privacy is formally characterized by $\epsilon$ and $\delta$ and denoted as $(\epsilon, \delta)-\text{DP}$. Lower $\epsilon$ and $\delta$ imply more privacy.  The standard procedure to train neural networks with DP is Differential Private SGD  (DPSGD, \citet{abadi2016deep}). DPSGD is a private variant of SGD in which per-sample parameter gradients are clipped, and Gaussian noise is added before the update step. The noise magnitude depends on $\epsilon, \delta,$ and model size and  drastically impacts utility~\cite{tramer2021differentially}.

Recently, \citet{yu2022differentially, li2022large} demonstrated that the utility of differential private finetuning is at par with non-private training. One of the key insights is that the parameter-efficient methods are better private learners than full finetuning. Intuitively, the amount of noise scales with parameters and fewer parameters implies less noise is added during training. Naturally, this encouraged us to evaluate \layershift\ and \latentprompt\ for private learning.
To this end, we use the same setup as \citet{yu2022differentially}. In particular, we consider the tasks with more than 10K samples in the GLUE benchmark and train to achieve $(\epsilon=6.7, \delta=10^{-6})-\text{DP}$. Different from \Cref{sec:seq_clf}, we report accuracy for all the tasks here.
We compare against the methods reported by \citet{yu2022differentially}, which include \lora, \adapters, and \compacters~\cite{karimi2021compacter}. Compacter is an improved and efficient version of the \adapters. RGP updates all the parameters, \ie, it is similar to full-finetuning but uses a different parametrization.

\paragraph*{Results.}
\Cref{tab:dp} reports the results of private finetuning \roberta\   under a fixed privacy budget ($\epsilon=6.7, \delta=10^{-6}$).
Due to using only a tiny number of parameters, the gap in the non-private and private utility of \layershift\ and \latentprompt\ is small.
Further, \layershift\ outperforms all the other methods on MNLI and QNLI tasks and is only second to the best (\lora) on QQP and SST-2 with \robertal. Similarly, \latentprompt\ and \layershift\ outperform all the other methods on the QNLI task with \robertab; however, \latentprompt's utility is lower on MNLI. \layershift's utility is generally comparable to other methods for all the tasks. Our approaches (\layershift\ and \latentprompt) may be more effective for larger models as those are  easier to tune with fewer parameters~\cite{lester-etal-2021-power}.

%% file: tables/glue_results.tex
\begin{table*}
  \centering
  \fontsize{9.2}{11}\selectfont
  \setlength{\tabcolsep}{3pt}

  \begin{tabular}[]{l r c c c c c c c c c}
    \toprule
    \multirow{2}{*}{Method}
    &\makecell[tc]{\#\\ Params}
    &\makecell[tc]{MNLI  \\ (392,702)}
    &\makecell[tc]{QQP   \\ (363,846)}
    &\makecell[tc]{QNLI  \\ (104,743)}
    &\makecell[tc]{SST-2 \\  (67,349)}
    &\makecell[tc]{CoLA  \\   (8,551)}
    &\makecell[tc]{STS-B \\   (5,749)}
    &\makecell[tc]{MRPC  \\   (3,668)}
    &\makecell[tc]{RTE   \\   (2,490)}
    &\multirow{2}{*}{Avg.}\\
    \cmidrule(r){1-1}
    \cmidrule(r){2-2}
    \cmidrule(lr){3-3}
    \cmidrule(lr){4-4}
    \cmidrule(lr){5-5}
    \cmidrule(lr){6-6}
    \cmidrule(lr){7-7}
    \cmidrule(lr){8-8}
    \cmidrule(lr){9-9}
    \cmidrule(lr){10-10}
    \cmidrule(lr){11-11}
    Finetuning                            & 355M  & 90.2 & 92.2 & 94.7 & 96.4 & 68.0 & 92.4 & 90.9 & 86.6 & 88.9 \\
    \adapters                             & 3M    & 90.4 & 88.5 & 94.7 & 96.3 & 67.4 & 92.5 & 92.9 & 83.4 & 88.3 \\
    \rowcolor{\tblrowcolor}
    Linear Classifier                     & 3.1K  & 70.9 & 77.1 & 78.8 & 89.8 & 48.9 & 73.8 & 83.8 & 72.2 & 74.4 \\
    \lora                                 & 800K  & 90.8 & 88.8 & 94.9 & 96.2 & 68.2 & 92.6 & 93.6 & 87.4 & 89.0 \\
    \cmidrule(lr){1-11}
    \rowcolor{\tblrowcolor}
    \protect{\warp\textsubscript{1}}      & 4.1K  & 83.9 & 81.6 & 87.6 & 93.8 & 46.1 & 80.4 & 84.7 & 72.6 & 78.8 \\
    \protect{\warp\textsubscript{8}}      & 11K   & 87.6 & 83.8 & 93.0 & 95.4 & 57.4 & 81.0 & 85.6 & 72.9 & 82.1 \\
    \protect{\warp\textsubscript{20}}     & 25K   & 88.2 & 84.5 & 93.5 & 96.0 & 60.6 & 88.6 & 90.8 & 75.8 & 84.8 \\
    \protect{\warp\textsubscript{MNLI}}   & 25K   &   -  &   -  &   -  &   -  &   -  & 91.0 & 91.2 & 86.3 & 86.4 \\
    \cmidrule(lr){1-11}
    \protect{\lora} \tiny[rank = 1]       & 101K  & 90.0 & 87.1 & 94.3 & 95.9 & 63.3 & 91.9 & 92.9 & 85.6 & 87.6 \\
    \protect{\adapters} \tiny[m = 1]      & 150K  & 90.4 & 88.0 & 94.7 & 95.9 & 68.0 & 92.1 & 92.6 & 85.6 & 88.4 \\
    \protect{\bitfit}                     & 276K  & 90.4 & 87.3 & 94.5 & 95.4 & 66.0 & 92.1 & 93.3 & 83.4 & 87.8 \\
    \cmidrule(lr){1-11}
    \rowcolor{\tblrowcolor}
    Ours \tiny[\zdim\ = 1,024]            & 4.1K  & \result{85.8}{0.23} & \result{83.2}{0.15} & \result{92.2}{0.24} & \result{94.7}{0.57} & \result{59.6}{2.43} & \result{90.4}{0.41} & \result{91.1}{0.56} & \result{81.5}{2.18} & 84.8 \\
    \rowcolor{\tblrowcolor}
    Ours \tiny[\zdim\ = 2,048]            & 5.1K  & \result{87.4}{0.08} & \result{84.1}{0.09} & \result{92.9}{0.28} & \result{94.9}{0.34} & \result{60.7}{2.11} & \result{90.7}{0.30} & \result{91.3}{0.84} & \result{83.5}{1.67} & 85.7 \\
    Ours \tiny[\zdim\ = 10K]              & 13.1K & \result{89.0}{0.14} & \result{85.5}{0.10} & \result{93.4}{0.19} & \result{95.2}{0.36} & \result{62.8}{1.43} & \result{91.5}{0.24} & \result{89.5}{4.17} & \result{84.1}{1.10} & 86.4 \\
    \protect{\latentprompt\textsubscript{1}}
    \tiny{[\zdim\ = 10K]}                 & 13.1K & \result{86.8}{1.26} & \result{84.2}{0.52} & \result{93.2}{0.20} & \result{95.3}{0.37} & \result{57.3}{2.61} & \result{89.1}{0.69} & \result{89.7}{1.41} & \result{79.6}{1.32} & 84.4 \\
    \cmidrule(lr){1-11}
    Ours \tiny{[\zdim\ = 24,576]} (max)   & 27.7K & 89.5 & 86.5 & 93.4 & 95.6 & 64.0 & 91.5 & 92.1 & 87.7 & 87.5 \\
    \bottomrule
  \end{tabular}
  \caption[]{
    Results of finetuning  \robertal\ with different methods on GLUE Development set. The bracketed numbers in the heading are training set sizes. \# Params are per-task trainable parameters. Rows with very few (< 10K) parameters are highlighted in \tblrowcolorname\ to facilitate comparison.
    Finetuning results are from \citet{liu2019roberta}, and
    \adapters\ (3M) and \warp\ results are from \citet{hambardzumyan-etal-2021-warp}.
    Linear results are the best of linear classifier and \warp\textsubscript{0} performance from \citet{hambardzumyan-etal-2021-warp}.\footnotemark\  \warp\textsubscript{MNLI} used an additional intermediate step of supervised training on the MNLI dataset.
    \lora\ (800K) results are adapted from \citet{hu2022lora}.\footnotemark\ The standard deviations are computed over 5 training runs with different seeds. Due to computational limitations, we report error bars for our methods only.}
  \label{tab:roberta_large_glue_result}
\end{table*}

%% file: tables/glue_results_footnote.tex
\addtocounter{footnote}{-1}
\protect{\footnotetext{\warp\textsubscript{0} feeds \mask\ representations to the classifier head, whereas the linear classifier uses \cls representations.}}
\addtocounter{footnote}{+1}
\protect{\footnotetext{Since they report different metrics, we evaluated \lora\ from the provided checkpoints on MNLI, STS-B, and QQP.}}

%% file: tables/glue_roberta_base.tex
\begin{table*}
    \centering
    \fontsize{9.2}{11}\selectfont
    \setlength{\tabcolsep}{3pt}

    \begin{tabular}[]{l r c c c c c c c c c}
        \toprule
        {Method}
        &\% params
        &MNLI
        &QQP
        &QNLI
        &SST-2
        &CoLA
        &STS-B
        &MRPC
        &RTE
        & {Avg.}\\
        \cmidrule(r){1-1}
        \cmidrule(l){2-2}
        \cmidrule(lr){3-3}
        \cmidrule(lr){4-4}
        \cmidrule(lr){5-5}
        \cmidrule(lr){6-6}
        \cmidrule(lr){7-7}
        \cmidrule(lr){8-8}
        \cmidrule(lr){9-9}
        \cmidrule(lr){10-10}
        \cmidrule(lr){11-11}
        Finetuning                          & 100\%    & 86.4 & 88.0 & 92.3 & 94.2 & 61.1 & 90.6 & 92.5 & 77.4 & 85.3 \\
        BitFit                              & 0.09\%   & 85.8 & 85.2 & 91.9 & 93.7 & 60.1 & 90.6 & 91.9 & 71.8 & 83.9 \\
        \protect{\lora} \tiny{[rank = 1]}   & 0.04\%   & 86.3 & 85.6 & 92.7 & 94.3 & 60.1 & 90.1 & 91.3 & 76.2 & 84.6 \\
        \protect{\adapters} \tiny{[m = 1]}  & 0.05\%   & 86.7 & 86.1 & 92.0 & 94.3 & 61.4 & 91.0 & 92.3 & 78.3 & 85.3 \\
        \cmidrule(lr){1-11}
        Ours \tiny{[\zdim\ = 1,024]}        & 0.003\%  & \result{80.6}{0.26} & \result{80.9}{0.09} & \result{89.1}{0.53} & \result{92.6}{0.27} & \result{55.5}{1.99} & \result{89.4}{0.19} & \result{90.4}{0.76} & \result{76.9}{1.87} & 81.9 \\
        Ours \tiny{[\zdim\ = 5K]}           & 0.007\%  & \result{83.6}{0.16} & \result{83.2}{0.11} & \result{90.6}{0.21} & \result{93.1}{0.45} & \result{59.1}{1.74} & \result{89.9}{0.28} & \result{90.7}{0.88} & \result{76.7}{1.84} & 83.4 \\
        \protect\latentprompt\textsubscript{1} 
        \tiny{[\zdim\ = 5K]}                  & 0.007\%  & \result{81.9}{0.78} & \result{81.6}{0.66} & \result{88.2}{1.24} & \result{92.5}{0.60} & \result{43.4}{9.12} & \result{86.3}{1.75} & \result{82.5}{3.45} & \result{69.5}{1.36} & 78.2 \\
        \cmidrule(lr){1-11}
        Ours \tiny{[\zdim\ = 9,216] (max)}        & 0.011\%  &  84.4 & 83.9 & 90.5 & 93.7 & 58.8 & 90.1 & 90.8 & 79.4  & 83.9\\
        \bottomrule
    \end{tabular}
    \caption{
    Results of finetuning \robertab\ with different methods on GLUE Development set.
    Finetuning results are taken from \citet{ben-zaken-etal-2022-bitfit}. \robertab\
    has 108 million parameters. The standard deviations are computed over 5 training runs with different seeds. Due to computational limitations, we report error bars for our methods only.
    }
    \label{tab:roberta_base_glue_result}
\end{table*}

%% file: tables/ner_results.tex
\begin{table}
    \centering
    \fontsize{10}{11}\selectfont
    \setlength{\tabcolsep}{4pt}
    \begin{tabular}{l r c c}
        \toprule
        Method
        &\# params
        &Test
        &Validation \\
        \cmidrule(r){1-1}
        \cmidrule(l){2-2}
        \cmidrule(lr){3-3}
        \cmidrule(lr){4-4}
        Finetuning                          & 108M    & 91.35 & 94.97 \\
        Linear Classifier                   & 7K      & 82.02 & 85.94 \\
        \protect{\lora} \tiny[rank = 1]     & 44K     & 89.50 & 93.38 \\
        \protect{\adapters} \tiny[m = 1]    & 63K     & 90.09 & 93.55 \\
        \bitfit                             & 109K    & 89.83 & 93.62 \\
        \warp\textsubscript{20}             & 22.3K   & 86.03 & 89.89 \\
        \cmidrule(lr){1-4}
        Ours \tiny[\zdim\ = 1,024]          & 8K      & 86.49 & 89.37 \\
        Ours \tiny[\zdim\  = 5K]            & 12K     & 88.30 & 91.38 \\
        \latentprompt\textsubscript{1} \tiny[\zdim\ = 5K]
                                            & 12K     & 87.08 & 90.93 \\
        \bottomrule
    \end{tabular}
    \caption{Results of finetuning \bertb\texttt{-cased}\ for NER task on CoNLL-2003 (English) dataset.}
    \label{tab:ner_result}
\end{table}

%% file: tables/profiling.tex
\begin{table*}
    \fontsize{10}{11}\selectfont
    \setlength{\tabcolsep}{4pt}
    \begin{subtable}[c]{0.45\textwidth}
        \centering
        \begin{tabular}{l c c }
        \toprule
            Method                             & Time (s) & Memory (GB) \\
            \cmidrule(r){1-1} \cmidrule(lr){2-2} \cmidrule(lr){3-3}
            Finetuning	                       & 3291	  & 15.6 \\
            \protect{\bitfit}	               & 2083	  & 8.6  \\
            \protect{\lora\ \tiny[rank = 1]}   & 2019     & 13.0 \\
            \protect{\adapters\ \tiny[m = 1]}  & 2289	  & 13.1 \\
            \protect{\warp\textsubscript{20}}  & 1869	  & 9.0  \\
            \protect{Ours \tiny[\zdim\ = 10K]} & 1764	  & 9.3  \\
        \bottomrule
        \end{tabular}
        \caption{\robertal}
    \end{subtable}
    \begin{subtable}[c]{0.45\textwidth}
        \centering
        \begin{tabular}{l c c}
        \toprule
            Method                              & Time (s) & Memory (GB)\\
            \cmidrule(r){1-1} \cmidrule(lr){2-2} \cmidrule(lr){3-3}
            Finetuning                          & 1227	   & 5.8 \\
            \protect{\bitfit}	                & 819	   & 3.3 \\
            \protect{\lora\ \tiny[rank = 1]}	& 1026	   & 4.9 \\
            \protect{\adapters\ \tiny[m = 1]}   & 1385	   & 4.8 \\
            \protect{\warp\textsubscript{20}}   & 635	   & 3.5 \\
            \protect{Ours \tiny[\zdim\ = 5K]}	& 558	   & 3.3 \\
        \bottomrule
        \end{tabular}
        \caption{\robertab}
        \label{tab:profile_roberta_base}
    \end{subtable}
    \caption{Memory and execution time for training 1 epoch on QNLI dataset (104,743 samples) with batch size 8. We report the maximum memory allocated during the training on a Quadro RTX 8000 GPU.   }
    \label{tab:profiling}
\end{table*}

%% file: tables/glue_results_pvt.tex
\newcommand{\best}[1]{\textbf{#1}}

\begin{table*}
    \fontsize{10}{11}\selectfont
    \setlength{\tabcolsep}{4pt}

    \begin{subtable}[c]{0.49\textwidth}
        \centering
        \begin{tabular}{l c c c c }
        \toprule
                                & MNLI         & QQP         & QNLI         & SST-2      \\
            \cmidrule(lr){2-2}
            \cmidrule(lr){3-3}
            \cmidrule(lr){4-4}
            \cmidrule(lr){5-5}
            \multicolumn{5}{c}{\highlight{Non-Private Training}}\\
            Finetuning                & 90.2         & 92.2        & 94.7         & 96.4        \\
            Ours \tiny[\zdim\ = 10K]  & 89.1         & 89.1        & 93.5         & 95.9        \\
            \latentprompt\textsubscript{1} \tiny[\zdim\ = 10K]
                                      & 89.0         & 88.9        & 93.5         & 95.5        \\
            \midrule
            \multicolumn{5}{c}{\highlight{Private Training}}\\
            Ours \tiny[\zdim\ = 10K]  & \best{88.0}  & 86.9        & \best{91.2}  & 94.5        \\
        \latentprompt\textsubscript{1} \tiny[\zdim\ = 10K]
                                & 87.7         & 86.3        & 91.1         & 94.4        \\
            RGP                 & 86.1         & 86.7        & 90.0         & 93.0        \\
            \adapters           & 87.7         & 86.3        & 90.7         & 93.9        \\
            \compacters         & 87.5         & 86.2        & 90.2         & 94.2        \\
            \lora               & 87.8         & \best{87.4} & 90.8         & \best{95.3} \\
        \bottomrule
        \end{tabular}
        \caption{Finetuning with \robertal}
        \label{tab:private_roberta_large}
    \end{subtable}
    \begin{subtable}[c]{0.49\textwidth}
        \centering
        \begin{tabular}{l c c c c }
        \toprule
                                & MNLI          & QQP         & QNLI        & SST-2     \\
            \cmidrule(lr){2-2}
            \cmidrule(lr){3-3}
            \cmidrule(lr){4-4}
            \cmidrule(lr){5-5}
            \multicolumn{5}{c}{\highlight{Non-Private Training}}\\
            Finetuning          & 87.6          & 91.9        & 92.8        & 94.8        \\
            Ours \tiny[\zdim\ = 5K]   & 83.6          & 87.4        & 90.8        & 93.7        \\
            \latentprompt\textsubscript{1} \tiny[\zdim\ = 5K]
                                & 83.4          & 87.2        & 90.7        & 93.3        \\
            \midrule
            \multicolumn{5}{c}{\highlight{Private Training}}\\
        Ours \tiny[\zdim\ = 5K]       & 83.0          & 84.9        & 87.6        & 92.4        \\
            \latentprompt\textsubscript{1} \tiny[\zdim\ = 5K]
                                & 81.3          & 84.7        & \best{87.9} & 92.0        \\
            RGP                 & 80.1          & 85.5        & 87.2        & 91.6        \\
            \adapters           & 83.4          & 85.6        & 87.5        & \best{92.5} \\
            \compacters         & 82.6          & 84.7        & 85.1        & 92.3        \\
            \lora               & \best{83.5}   & \best{85.7} & 87.3        & 92.2        \\
        \bottomrule
        \end{tabular}
        \caption{Finetuning with \robertab}
        \label{tab:private_roberta_base}
    \end{subtable}
    \caption{Results of differential private finetuning on GLUE Development set. Non-private finetuning and Private training results for RGP, Compacter, Adapter, and LoRA are from \citet{yu2022differentially}. Private models were trained to achieve $\epsilon=6.7$ for all datasets and  $\delta = 10^{-6}$ for MNLI, QQP, and QNLI and  $\delta = 10^{-5}$ for SST-2. For our method, privacy parameters are $\epsilon=6.7$ and $\delta=10^{-6}$ for all datasets (\ie, identical or stricter than the baselines).}
    \label{tab:dp}
\end{table*}

%% file: sections/related.tex
\seperator
\section{Related Work}\label{sec:related}
Prompt tuning and task-specific finetuning are standard ways to prime LMs for downstream tasks~\cite{liu2021pre,treviso2022efficient}. Prompt tuning inserts task-specific information or parameters around the input. Various versions exist, such as manual prompt-tuning, discrete prompt search~\cite{shin-etal-2020-autoprompt}, and continuous search~\cite{hambardzumyan-etal-2021-warp}. Prompt tuning is highly parameter efficient but is generally only effective for larger LMs~\cite{lester-etal-2021-power, yang2022prompt}. Due to joint reparametrization, our method uses a similar number of parameters as prompt-tuning methods but outperforms them.

Several parameter-efficient LM finetuning methods have been proposed, such as \adapters~\cite{houlsby2019parameter}, \lora~\cite{hu2022lora}, Prefix-Tuning~\cite{li-liang-2021-prefix}, and Parallel Adapters~\cite{he2022towards}.  Further improvements try to maintain the utility while reducing the parameters such as Compacter~\cite{karimi2021compacter} that parameterizes weight matrices via the sum of Kronecker products, pruning adapter layers~\cite{ruckle-etal-2021-adapterdrop, pfeiffer-etal-2021-adapterfusion} and gating mechanisms to choose the best modules~\cite{mao-etal-2022-unipelt}. These methods outperform prompt tuning but use more parameters. In contrast, we outperform prompt tuning while using similar number of parameters and are competitive with other finetuning approaches.

Our approach could be of independent interest for understanding intriguing properties of pretrained language models, the role of different parameters, and sharing parameters across layers. \citet{ben-zaken-etal-2022-bitfit,cai2020tinytl} have shown that pretrained models can be finetuned by only updating the bias parameters, but unlike us, they do not share parameters. \citet{gheini-etal-2021-cross} finetune only cross attention layers for machine translation. \citet{zhou-etal-2022-making} share only output layers across tasks, but parameters across different layers are not shared. \citet{zhou2022efficiently} have shown that task embeddings can be derived from task-specific finetuned parameters. The \zvector\ in our approach can also be helpful as a task-embedding.

Parameters derived by fixed random transformations a few parameters have previously been used to study the task's intrinsic dimensionality~\cite{li2018measuring,aghajanyan-etal-2021-intrinsic}. Those works focus on weight matrices. While insightful, these are cumbersome to train for real-world deployment.   Instead, we  focus on bias or embeddings, providing a tractable operationalization for regular training and finetuning while using similar order of parameter count. For example, \citet{aghajanyan-etal-2021-intrinsic} show that  the intrinsic dimension of the QQP dataset with \robertal\ is 774, \ie, at least 774 parameters are required to achieve within 90\% of full finetuning performance. \layershift\ achieves an F1-score of 83.2, more than  90\%  of full finetuning performance on QQP with 4.1K parameters ($92.2\times0.9=83.0$).

%% file: sections/conclusion.tex
\section{Conclusion}\label{sec:conclusion}
We introduce a multilayer LM finetuning technique where task-specific parameters are derived from a single vector. We show two instantiations of this technique --- \layershift\ and \latentprompt. \layershift\ introduced shifts in the output activation of each transformer block, whereas \latentprompt\ inserted prompts in each transformer block. These methods require only a tiny fraction  of the original language model parameters (similar to prompt-tuning) and outperform previous methods that use a similar number of per-task parameters. Despite the drastic reduction in the number of parameters, we demonstrate that these perform just as well as full finetuning for sentence and token classification tasks (only at max a 5\% difference in performance). The high parameter efficiency leads to better training speed and resource utilization and improves private training.

%% file: sections/limits.tex
\section{Limitations}\label{sec:limitations}
\paragraph*{Experiments.}
In this work, we propose new methods for finetuning language models. We acknowledge that similar to previous approaches, our experiments are limited to English datasets and specific supervised tasks. However, our method does not use language- or task-specific tricks and should apply to other languages and tasks.

\paragraph*{Method.}
As demonstrated in \Cref{sec:experiments}, \layershift\ is computationally efficient and performs comparably to the full finetuning for small datasets. Moreover, its parameter and memory efficiency makes it an excellent private learner.  However, it may underperform by a few points compared to full-finetuning larger datasets with higher intrinsic dimensionality due to using very few parameters. For example, \layershift\ struggles with generative tasks such as text summarization, as generative tasks are more complex and involve making predictions over the whole vocabulary. In contrast, classification tasks have relatively fewer output labels.  In our initial experiments, \layershift\ reached a ROUGE-2 score of 12.93 on the XSum summarization task~\cite{narayan-etal-2018-dont} with pretrained \texttt{BART}, whereas full finetuning achieves a  score of 21.94~\cite{he2022towards}.

The limitations of \layershift\ are due to the small number of parameters it updates. Since shift is applied to only certain biases, the number of parameters can not be increased beyond a limit. However, we show that \layershift\ is a more efficient and performant alternative to the methods that use a similar number of per-task parameters.
Moreover, we showed that joint reparametrization improves parameter efficiency of other methods. As such, this principle can be extended to methods that are not restricted by a maximum limit on the number of parameters.  For example, \latentprompt's parameters can be naturally increased by increasing the prompt length, which should improve the results further (details in \Cref{sec:latent_prompt}).

%% file: sections/ethics.tex
\section{Ethics Statement}\label{sec:ethics}
We propose a parameter-efficient method to tune transformer-based language models. The ethical implications are similar to the finetuning methods proposed before us. Our method improves parameter and computational efficiency, which should have an overall positive impact by reducing costs and enabling low-resource applications. Further, the positive private training results should encourage its adoption in real-world setups.

%% file: appendix/main.tex
\cleardoublepage
\appendix

\twocolumn[
\begin{center}
    {\large \bf Supplementary: \papertitle}\newline
\end{center}
]
\input{appendix/tables/pos_ablation}
\input{appendix/latent_prompt.tex}

\section{Ablations}\label{sec:ablations}

    Here we evaluate alternative hyperparameter choices for \layershift\ by performing ablation studies concerning the position of shifts, initialization of parameters, and using \mask\ \vs \cls\ representations. Overall, our results are relatively less sensitive to these choices.

    \input{appendix/tables/init_ablation}

\input{appendix/pos_ablation}
    \input{appendix/init_ablation}

    \input{appendix/tables/mask_vs_cls}
    \input{appendix/mask_vs_cls}

\input{appendix/hyperparams}

%% file: appendix/tables/pos_ablation.tex
\begin{table*}
    \centering
    \fontsize{10}{11}\selectfont
    \setlength{\tabcolsep}{4pt}
    \begin{tabular}{l l c c c c c c c c c}
        \toprule
        & Position
        & MNLI
        & QQP
        & QNLI
        & SST-2
        & CoLA
        & STS-B
        & MRPC
        & RTE
        & Avg.\\
        \cmidrule(lr){2-2}
        \cmidrule(lr){3-3}
        \cmidrule(lr){4-4}
        \cmidrule(lr){5-5}
        \cmidrule(lr){6-6}
        \cmidrule(lr){7-7}
        \cmidrule(lr){8-8}
        \cmidrule(lr){9-9}
        \cmidrule(lr){10-10}
        \cmidrule(lr){11-11}
        \zdim\ = 1,024 & \texttt{attention}      & 80.3 & 81.0 & 88.7 & 93.2 & 57.9 & 89.5 & 91.1 & 73.6 & 81.93 \\
        \zdim\ = 1,024 & \texttt{intermediate}   & 80.0 & 81.2 & 88.9 & 93.2 & 59.6 & 89.7 & 92.3 & 76.2 & 82.64 \\
        \zdim\ = 1,024 & \texttt{output}         & 80.4 & 80.9 & 89.3 & 93.1 & 59.5 & 89.3 & 91.7 & 77.6 & 82.72 \\
        \midrule
        \zdim\ = 5K & \texttt{intermediate}   & 83.7 & 83.7 & 90.2 & 93.2 & 58.4 & 89.9 & 92.1 & 78.0 & 83.65 \\
        \zdim\ = 5K & \texttt{output}         & 83.4 & 83.4 & 90.6 & 93.2 & 59.3 & 90.4 & 91.9 & 77.6 & 83.74 \\
        \bottomrule
    \end{tabular}
    \caption{Effect of adding shifts at different position on sequence classification tasks (GLUE Development set) with \robertab\ as the pretrained model. All the results are with \cls\ representations.}
    \label{tab:pos_ablation}
\end{table*}

%% file: appendix/latent_prompt.tex
\section{\latentprompt: Improved Prompt Tuning}\label{sec:latent_prompt}

\input{figures/jr_warp.tex}

\Cref{fig:jr-warp-approach} summarizes \latentprompt\ with prompt length 1.  We introduce prompts or embeddings in each transformer block, similar to \citet{liu-etal-2022-p}. However, in our case, the prompts are reparametrized as random projections of a single vector ${\color{red}\zvector}\in \real^d$.\footnote{This reparametrization differs from the generally suggested reparametrization of using an MLP encoder to transform the prompts.}  The prompt is appended to the token embeddings for the first layer, \ie, the embedding layer.  Previous multi-layer prompt tuning approaches discard the transformed prompt from the previous layers and insert a new prompt at each layer~\cite{lester-etal-2021-power,liu-etal-2022-p}.
Instead, from the second transformer block onwards, we do not discard previous representations and add the prompt to the resulting representation (or the transformed prompt) from the previous layer.
${\color{ForestGreen}\randomW_\layervariable}$ and ${\color{red}\zvector}$ are initialized similarly to \layershift.

\warp\ appends prompt only to the token embeddings, and in \Cref{fig:jr-warp-approach}, this can be achieved by keeping only the lower arm  emitting from \zvector\ block and  setting $\randomW_0$ as the identity matrix. \Cref{fig:jr-warp-approach} shows prompt length 1, but it can be extended  to prompts longer than  length 1. However, our main aim is to evaluate performance while using parameters similar to \warp.
Therefore, we keep the prompt length to 1, and $\zdim$  is 10K and 5K in our experiments. When extending the prompt length to more than one, there are multiple ways to reparametrize prompts. For example, reparametrize prompts within the same layer from a single \zvector\ or reparametrize prompts within the same index or time step from a single \zvector, as we have done in this work.

%% file: figures/jr_warp.tex
\begin{figure}
    \centering
    {
        \includegraphics[width=0.45\textwidth,clip, trim={17 0 7 0 }]{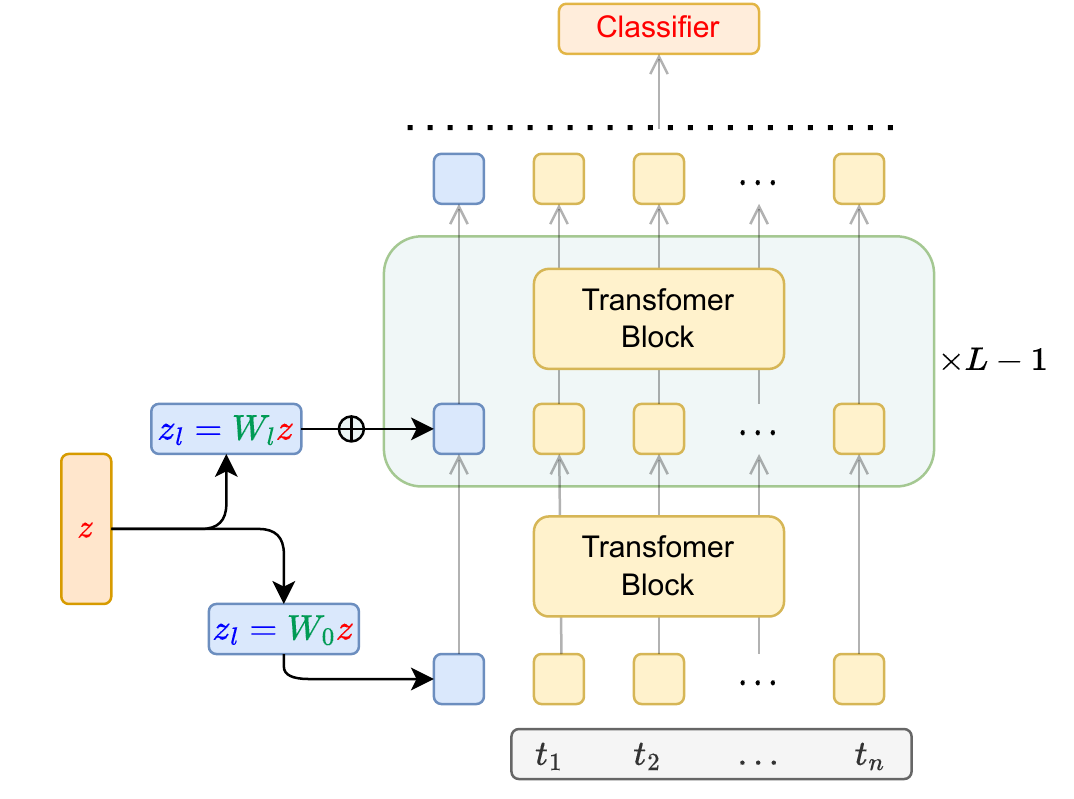}
    }
    \caption{\latentprompt\ for sequence classification tasks with prompt length 1.
    Similar to \Cref{fig:approach}, trainable modules are indicated in red.  $[t_i]_{i=1}^n$ denotes the sequence of tokens. The prompts are obtained by the projections
    \protect{${\color{ForestGreen}\randomW_\layervariable}{\color{red}\zvector}$}. For the first layer, the prompt is appended to the token embeddings. We add the prompt to the transformed prompt from the previous block for the rest of the layers. ${\color{red}\zvector}$ is shared across all the blocks. }
    \label{fig:jr-warp-approach}
\end{figure}

%% file: appendix/tables/init_ablation.tex
\begin{table*}
    \centering
    \fontsize{10}{11}\selectfont
    \setlength{\tabcolsep}{4pt}
    \begin{tabular}{l c c c c c c}
        \toprule
        Initialization
        & SST-2
        & CoLA
        & STS-B
        & MRPC
        & RTE
        & Avg.\\
        \cmidrule(l){1-1}
        \cmidrule(lr){2-2}
        \cmidrule(lr){3-3}
        \cmidrule(lr){4-4}
        \cmidrule(lr){5-5}
        \cmidrule(lr){6-6}
        \cmidrule(lr){7-7}
        $\zvector, \randomW_\layervariable \in \{\normal , \uniform\}$
            & 95.0 & 63.6 & 90.8 & 92.1 & 84.8 &  85.27\\
        $\zvector = \mathbf 0$, $\randomW_\layervariable \in \{\normal, \uniform\}$
            & 95.2 & 66.0 & 90.4 & 91.7 & 83.8 & 85.42\\
        $\zvector \in \{\normal , \uniform\}$, $\randomW_\layervariable= \mathbf I$
            & 95.1 & 62.7 & 90.4 & 92.6 & 83.8 & 84.92\\
        \bottomrule
    \end{tabular}
    \caption{Effect of different initialization of \layershift\ parameters on sequence classification tasks (GLUE Development set) with
    \robertal\ as the pretrained model. All the results use \mask\ representations and \zdim\ = 1,024.}
    \label{tab:init_ablation}
\end{table*}

%% file: appendix/pos_ablation.tex
\subsection{Adding shifts to other activations}\label{sec:pos_ablations}

In the main paper, we showed the results of adding shifts, \ie, random projections from a trainable vector to the output layer's activation. These shifts can also be added to other activations, such as the activations after attention layers, intermediate feed-forward layers, or a combination of these.  We evaluate these choices on tasks from the GLUE benchmark, and \Cref{tab:pos_ablation} summarizes our findings.

We find that the performance of shifting attention activations is similar to shifting output activations in most cases except for RTE and CoLA. Similar observations hold for intermediate activations. Shifting activations from intermediate feed-forward layers performed similarly for all tasks compared to output activations. These observations do not change  when we increase the trainable parameters. Shifting output activations performed slightly better in terms of average performance computed across all tasks. Moreover, the intermediate activations have a higher dimension than the output activation (3,072 \vs 768 for \robertab). Therefore, intermediate activations required maintaining bigger random projection matrices ($\randomW_\layervariable$) during training.

In summary, other choices can perform similarly. We chose output activations due to their smaller dimension and transformers using layer norm immediately after it, which can take care of sudden drifts in activations, etc.

%% file: appendix/init_ablation.tex
\subsection{Initialization}\label{sec:init_ablations}

Regarding the initialization of \zvector\ and $\randomW_\layervariable$, we have several choices.
\zvector\ can be initialized randomly or with all zeros. Like \citet{hambardzumyan-etal-2021-warp}, we report results with random initialization for \zvector\ in the main paper. In particular, it is initialized as $\normal(0, \sigma=\frac 1 {\sqrt \zdim})$ or $\uniform(-\frac 1 {\sqrt {12 \zdim}},\frac 1 {\sqrt {12 \zdim}} )$. The projection matrices, $\randomW_\layervariable$, are also initialized randomly with identical distributions as \zvector. With these initialization choices, the variance of $\zvector_\layervariable$ is $\frac 1 d$ in each dimension.
We consider the choice of Gaussian or Uniform initialization as a hyperparameter.

\Cref{tab:init_ablation} shows the effect of different initialization on  performance for sequence classification tasks. The results are relatively less sensitive to initialization.  When both \zvector\ and weight matrices are randomly initialized, the performance is better on STS-B, MRPC, and RTE than when \zvector\ is initialized as all zeros. However, the average performance of all zeros is higher due to its performance being much higher on CoLA.

For the particular case of \zdim\ = 1024, \ie, the dimension of \zvector\ is the same as the activations, we can initialize $\randomW_\layervariable$ as identity. In this case, all the blocks are shifted with the same vector. This performed similarly or worse on all tasks except MRPC. Random projections allow the model to select different parts of \zvector\ for each transformer block. The above-mentioned result partly demonstrates the utility of using random projection matrices.

%% file: appendix/tables/mask_vs_cls.tex
\begin{table*}
    \centering
    \fontsize{10}{11}\selectfont
    \setlength{\tabcolsep}{4pt}
    \begin{tabular}[]{l c c c c c c c c c}
        \toprule
        Model
        & MNLI
        & QQP
        & QNLI
        & SST-2
        & CoLA
        & STS-B
        & MRPC
        & RTE
        & Avg.\\
        \cmidrule(l){1-1}
        \cmidrule(lr){2-2}
        \cmidrule(lr){3-3}
        \cmidrule(lr){4-4}
        \cmidrule(lr){5-5}
        \cmidrule(lr){6-6}
        \cmidrule(lr){7-7}
        \cmidrule(lr){8-8}
        \cmidrule(lr){9-9}
        \cmidrule(lr){10-10}
        \multicolumn{10}{c}{\underline{\robertab}}\\
        \zdim\ = 1,024, \mask & 80.8 & 80.9 & 89.8 & 92.9 & 57.6 & 89.5 & 91.0 & 78.7 & 82.65 \\
        \zdim\ = 1,024, \cls  & 80.4 & 80.9 & 89.3 & 93.1 & 59.5 & 89.3 & 91.7 & 77.6 & 82.72 \\
        \zdim\ = 5K, \mask    & 83.6 & 83.2 & 90.8 & 93.7 & 61.3 & 90.3 & 91.3 & 79.4 & 84.21  \\
        \zdim\ = 5K, \cls     & 83.4 & 83.4 & 90.6 & 93.2 & 59.3 & 90.4 & 91.9 & 77.6 & 83.74 \\
        \midrule
        \multicolumn{10}{c}{\underline{\robertal}}\\
        \zdim\ = 1024, \mask & 86.2 & 83.3 & 92.2 & 95.0 & 63.6 & 90.8 & 92.1 & 84.8 &	86.01 \\
        \zdim\ = 1024, \cls  & 86.3 & 83.1 & 92.3 & 95.1 & 61.6 & 90.5 & 92.6 & 82.3 &	85.47 \\
        \zdim\ = 10K, \mask  & 89.1 & 85.6 & 93.6 & 95.9 &	65.5 & 91.8	& 91.8 & 85.6 &	87.33 \\
        \zdim\ = 10K, \cls   & 89.1 & 85.7 & 93.6 & 95.8 &	64.0 & 91.7	& 91.8 & 86.6 &	87.29 \\
        \bottomrule
    \end{tabular}
    \caption{Comparing \layershift\ with \mask\ and \cls\ token representation on sequence classification tasks (GLUE Development set).}
    \label{tab:mask_vs_cls}
\end{table*}

%% file: appendix/mask_vs_cls.tex
\subsection{\mask\ \vs \cls\ Representations}\label{sec:mask_vs_cls}

As discussed in \Cref{sec:seq_clf}, we can use \cls\ or \mask\ representation for classification tasks. \Cref{tab:mask_vs_cls} compares this  with \robertab\ and \robertal\ models. In terms of average performance, we find that \mask\ token representations are better or similar to \cls\ token representations.

The choice of representations  mattered very little for bigger datasets (>10K samples), with the performance being similar for both choices. For smaller datasets, however, we do not see any clear patterns. On average, \mask\ token representation performed slightly better than \cls\ representation, echoing the observation of \citet{hambardzumyan-etal-2021-warp}. So we use \mask\ representation for all the results in the main paper.

%% file: appendix/hyperparams.tex
\section{Hyperparameters}\label{sec:hyperparams}

\input{appendix/tables/seq_clf_large.tex}

\input{appendix/tables/seq_clf_base.tex}

\input{appendix/tables/seq_clf_jr_warp.tex}

\input{appendix/tables/pvt_training.tex}

\input{appendix/tables/pvt_training_jr_warp.tex}

Our implementation is based on the Hugging Face Transformers library~\cite{wolf-etal-2020-transformers} and PyTorch 1.10 and 1.13.
We use AdapterHub~\cite{pfeiffer-etal-2020-adapterhub} for training \lora\ and \adapters\ models. We use PyTorch-Opacus~\cite{opacus} for private training. We mainly vary the learning rate and training epochs for all the methods. For \layershift\ and \latentprompt, we consider one additional hyperparameter --- Gaussian or Uniform initialization and disable all the dropout layers.

We use a similar training setup for \highlight{sequence classification} tasks as \citet{hambardzumyan-etal-2021-warp}. We tune the learning rate in $\{1e^{-4}, 3e^{-4}, 1e^{-3}, 3e^{-3}, 1e^{-2}, 3e^{-2} \}$ and use a linear learning rate scheduler with a warmup ratio of $0.06$. We train for $10$ or $20$ epochs with a batch size of $8$, and the gradient magnitudes are clipped to 1.0. \Cref{tab:hparams_seq_clf_base,tab:hparams_seq_clf_large} and \Cref{tab:hparams_seq_clf_jr_warp} list the best hyperparameters for each task for \layershift\ and \latentprompt, respectively. We find the best hyperparameters based on the performance on the validation set from a single training run. Then to report the error bars (in the main paper), we train several models with those best-found hyperparameters but with different random seeds.

For \highlight{token classification} tasks, we tune the learning rate in $\{1e^{-4}, 3e^{-4}, 1e^{-3}, 3e^{-3}, 1e^{-2}, 3e^{-2} \}$ with a linear learning rate scheduler and use a warmup ratio of $0.1$. We train for 5 epochs  with a batch size of 32. The best result for \layershift\ is obtained with uniform initialization and a learning rate of 0.01. The best result for \latentprompt\ is obtained with normal initialization and a 0.03 learning rate.

For \highlight{private training}, we replicated the setup of \citet{yu2022differentially} as much as possible. In particular, we tune the learning rate in $\{1e^{-3},  3e^{-3}, 1e^{-2}\}$ without any scheduler and train for $20$ epochs. We used a batch size of 2048 and  considered two per sample gradient clipping thresholds --- $0.1$ and $1.0$. We use the PRV accountant of \citet{gopi2021numerical} for privacy accounting, the same as \citet{yu2022differentially},  to keep the results comparable. Based on this accountant, the Gaussian noise magnitudes for MNLI, QQP, QNLI, and SST-2 were 0.643, 0.651, 0.831, and 0.925. \Cref{tab:hparams_pvt} and \Cref{tab:hparams_pvt_jr_warp} list the best hyperparameters for \layershift\ and \latentprompt.

%% file: appendix/tables/seq_clf_large.tex
\begin{table*}
    \centering
    \fontsize{10}{11}\selectfont
    \setlength{\tabcolsep}{4pt}

    \begin{tabular}{l  c c c c c c c c c}
    \toprule
        \multirow{2}{*}{Task}
        & \multicolumn{3}{c}{\zdim\ = 1,024 }
        & \multicolumn{3}{c}{\zdim\ = 2,048 }
        & \multicolumn{3}{c}{\zdim\ = 10K }\\
    \cmidrule(lr){2-4} \cmidrule(lr){5-7} \cmidrule(lr){8-10}
        & Initialization & LR & \# Epoch
        & Initialization & LR & \# Epoch
        & Initialization & LR & \# Epoch \\
    \cmidrule(l){1-1}
    \cmidrule(lr){2-2}
    \cmidrule(lr){3-3}
    \cmidrule(lr){4-4}
    \cmidrule(lr){5-5}
    \cmidrule(lr){6-6}
    \cmidrule(lr){7-7}
    \cmidrule(lr){8-8}
    \cmidrule(lr){9-9}
    \cmidrule(lr){10-10}
        RTE     &\normal  & $3e^{-2}$  & 10 & \uniform & $1e^{-2}$  & 20 & \uniform & $1e^{-2}$ & 10 \\
        MRPC    &\uniform & $1e^{-2}$  & 20 & \uniform & $1e^{-2}$  & 10 & \uniform & $1e^{-2}$ & 20 \\
        STSB    &\uniform & $3e^{-3}$  & 10 & \uniform & $1e^{-2}$  & 10 & \uniform & $3e^{-3}$ & 10 \\
        CoLA    &\normal  & $3e^{-3}$  & 20 & \uniform & $1e^{-2}$  & 10 & \normal  & $1e^{-2}$ & 10 \\
        SST-2   &\normal  & $3e^{-3}$  & 10 & \normal  & $1e^{-3}$  & 20 & \normal  & $3e^{-3}$ & 10 \\
        QNLI    &\uniform & $3e^{-3}$  & 20 & \normal  & $3e^{-3}$  & 20 & \uniform & $1e^{-3}$ & 10 \\
        QQP     &\uniform & $3e^{-3}$  & 20 & \normal  & $1e^{-3}$  & 20 & \normal  & $1e^{-3}$ & 20 \\
        MNLI    &\normal  & $3e^{-4}$  & 10 & \normal  & $1e^{-3}$  & 20 & \uniform & $1e^{-3}$ & 20 \\
    \bottomrule
    \end{tabular}
    \caption{Hyperparameters of best-performing \layershift\ models for sequence classification with \robertal. Results shown in \Cref{tab:roberta_large_glue_result}.}
    \label{tab:hparams_seq_clf_large}
\end{table*}

%% file: appendix/tables/seq_clf_base.tex
\begin{table*}
    \centering
    \fontsize{10}{11}\selectfont
    \setlength{\tabcolsep}{4pt}
    \begin{tabular}{l  c c c c c c}
    \toprule
        \multirow{2}{*}{Task} &
        \multicolumn{3}{c}{\zdim\ = 1,024 } &
        \multicolumn{3}{c}{\zdim\ = 5K }  \\
    \cmidrule(lr){2-4} \cmidrule(lr){5-7}
        & Initialization & LR &\# Epoch
        & Initialization & LR &\# Epoch     \\
    \cmidrule(l){1-1}
    \cmidrule(lr){2-2}
    \cmidrule(lr){3-3}
    \cmidrule(lr){4-4}
    \cmidrule(lr){5-5}
    \cmidrule(lr){6-6}
    \cmidrule(lr){7-7}
        RTE     & \uniform  &   $1e^{-2}$    & 20 & \uniform & $1e^{-2}$  &   10 \\
        MRPC    & \normal   &   $3e^{-2}$    & 10 & \normal  & $1e^{-2}$  &   10 \\
        STSB    & \normal   &   $1e^{-2}$    & 10 & \normal  & $1e^{-2}$  &   20 \\
        CoLA    & \uniform  &   $3e^{-3}$    & 10 & \normal  & $1e^{-2}$  &   10 \\
        SST-2   & \normal   &   $1e^{-3}$    & 10 & \uniform & $1e^{-2}$  &   10 \\
        QNLI    & \uniform  &   $3e^{-3}$    & 20 & \uniform & $3e^{-3}$  &   20 \\
        QQP     & \normal   &   $1e^{-3}$    & 20 & \normal  & $3e^{-3}$  &   20 \\
        MNLI    & \normal   &   $1e^{-3}$    & 10 & \normal  & $1e^{-3}$  &   20 \\
    \bottomrule
    \end{tabular}
    \caption{Hyperparameters of best-performing \layershift\  models for sequence classification with \robertab. Results shown in \Cref{tab:roberta_base_glue_result}.}
    \label{tab:hparams_seq_clf_base}
\end{table*}

%% file: appendix/tables/seq_clf_jr_warp.tex
\begin{table*}
    \centering
    \fontsize{10}{11}\selectfont
    \setlength{\tabcolsep}{4pt}

    \begin{tabular}{l  c c c c c c}
    \toprule
        \multirow{2}{*}{Task}
        & \multicolumn{3}{c}{\robertab\ (\zdim\ = 5K) }
        & \multicolumn{3}{c}{\robertal\  (\zdim\ = 10K) } \\
    \cmidrule(lr){2-4} \cmidrule(lr){5-7}
        & Initialization & LR & \# Epoch
        & Initialization & LR & \# Epoch \\
    \cmidrule(l){1-1}
    \cmidrule(lr){2-2}
    \cmidrule(lr){3-3}
    \cmidrule(lr){4-4}
    \cmidrule(lr){5-5}
    \cmidrule(lr){6-6}
    \cmidrule(lr){7-7}
        RTE     &\normal  & $1e^{-2}$  & 10 & \uniform & $1e^{-2}$  & 20 \\
        MRPC    &\uniform & $3e^{-3}$  & 10 & \normal  & $1e^{-2}$  & 10 \\
        STSB    &\normal  & $1e^{-2}$  & 20 & \normal  & $1e^{-2}$  & 20 \\
        CoLA    &\normal  & $1e^{-2}$  & 10 & \normal  & $1e^{-2}$  & 20 \\
        SST-2   &\uniform & $3e^{-3}$  & 10 & \uniform & $1e^{-2}$  & 20 \\
        QNLI    &\uniform & $3e^{-3}$  & 20 & \normal  & $1e^{-2}$  & 20 \\
        QQP     &\uniform & $3e^{-3}$  & 20 & \normal  & $3e^{-3}$  & 20 \\
        MNLI    &\normal  & $3e^{-3}$  & 20 & \normal  & $1e^{-3}$  & 20 \\
    \bottomrule
    \end{tabular}
    \caption{Hyperparameters of best-performing \latentprompt\ models for sequence classification with \roberta. Results shown in \Cref{tab:roberta_large_glue_result,tab:roberta_base_glue_result}.}
    \label{tab:hparams_seq_clf_jr_warp}
\end{table*}

%% file: appendix/tables/pvt_training.tex
\begin{table*}
    \centering
    \fontsize{10}{11}\selectfont
    \setlength{\tabcolsep}{4pt}

    \begin{tabular}{l  c c c c c c }
    \toprule
        \multirow{2}{*}{Task}
        & \multicolumn{3}{c}{\robertab}
        & \multicolumn{3}{c}{\robertal} \\
    \cmidrule(lr){2-4} \cmidrule(lr){5-7}
        & Initialization & LR & Grad. Clip Threshold
        & Initialization & LR & Grad. Clip Threshold \\
    \cmidrule(l){1-1}
    \cmidrule(lr){2-2}
    \cmidrule(lr){3-3}
    \cmidrule(lr){4-4}
    \cmidrule(lr){5-5}
    \cmidrule(lr){6-6}
    \cmidrule(lr){7-7}
        SST-2 &  \uniform  & $3e^{-3}$  & 0.1  & \uniform &  $1e^{-3}$ & 1.0 \\
        QNLI  &  \uniform  & $1e^{-2}$  & 1.0  & \normal  &  $1e^{-2}$ & 0.1 \\
        QQP   &  \normal   & $3e^{-3}$  & 1.0  & \normal  &  $3e^{-3}$ & 1.0 \\
        MNLI  &  \uniform  & $3e^{-3}$  & 1.0  & \uniform &  $3e^{-3}$ & 1.0 \\
    \bottomrule
    \end{tabular}
    \caption{Hyperparameters of best-performing \layershift\ models for private training. Results shown in \Cref{tab:dp}.}
    \label{tab:hparams_pvt}
\end{table*}

%% file: appendix/tables/pvt_training_jr_warp.tex
\begin{table*}
    \centering
    \fontsize{10}{11}\selectfont
    \setlength{\tabcolsep}{4pt}

    \begin{tabular}{l  c c c c c c }
    \toprule
        \multirow{2}{*}{Task}
        & \multicolumn{3}{c}{\robertab}
        & \multicolumn{3}{c}{\robertal} \\
    \cmidrule(lr){2-4} \cmidrule(lr){5-7}
        & Initialization & LR & Grad. Clip Threshold
        & Initialization & LR & Grad. Clip Threshold \\
    \cmidrule(l){1-1}
    \cmidrule(lr){2-2}
    \cmidrule(lr){3-3}
    \cmidrule(lr){4-4}
    \cmidrule(lr){5-5}
    \cmidrule(lr){6-6}
    \cmidrule(lr){7-7}
        SST-2 &  \normal   & $1e^{-2}$  & 0.1  & \normal  &  $1e^{-2}$ & 1.0 \\
        QNLI  &  \uniform  & $1e^{-2}$  & 1.0  & \normal  &  $1e^{-2}$ & 1.0 \\
        QQP   &  \normal   & $1e^{-2}$  & 1.0  & \uniform &  $1e^{-2}$ & 1.0 \\
        MNLI  &  \uniform  & $1e^{-2}$  & 1.0  & \uniform &  $1e^{-2}$ & 0.1 \\
    \bottomrule
    \end{tabular}
    \caption{Hyperparameters of best-performing \latentprompt\ models for private training. Results shown in \Cref{tab:dp}.}
    \label{tab:hparams_pvt_jr_warp}
\end{table*}

%% file: main.bbl
\begin{thebibliography}{43}
\expandafter\ifx\csname natexlab\endcsname\relax\def\natexlab#1{#1}\fi

\bibitem[{Abadi et~al.(2016)Abadi, Chu, Goodfellow, McMahan, Mironov, Talwar,
  and Zhang}]{abadi2016deep}
Martin Abadi, Andy Chu, Ian Goodfellow, H.~Brendan McMahan, Ilya Mironov, Kunal
  Talwar, and Li~Zhang. 2016.
\newblock \href {https://doi.org/10.1145/2976749.2978318} {Deep learning with
  differential privacy}.
\newblock In \emph{Proceedings of the 2016 ACM SIGSAC Conference on Computer
  and Communications Security}, CCS '16, page 308–318, New York, NY, USA.
  Association for Computing Machinery.

\bibitem[{Aghajanyan et~al.(2021)Aghajanyan, Gupta, and
  Zettlemoyer}]{aghajanyan-etal-2021-intrinsic}
Armen Aghajanyan, Sonal Gupta, and Luke Zettlemoyer. 2021.
\newblock \href {https://doi.org/10.18653/v1/2021.acl-long.568} {Intrinsic
  dimensionality explains the effectiveness of language model fine-tuning}.
\newblock In \emph{Proceedings of the 59th Annual Meeting of the Association
  for Computational Linguistics and the 11th International Joint Conference on
  Natural Language Processing (Volume 1: Long Papers)}, pages 7319--7328,
  Online. Association for Computational Linguistics.

\bibitem[{Ben~Zaken et~al.(2022)Ben~Zaken, Goldberg, and
  Ravfogel}]{ben-zaken-etal-2022-bitfit}
Elad Ben~Zaken, Yoav Goldberg, and Shauli Ravfogel. 2022.
\newblock \href {https://doi.org/10.18653/v1/2022.acl-short.1} {{B}it{F}it:
  Simple parameter-efficient fine-tuning for transformer-based masked
  language-models}.
\newblock In \emph{Proceedings of the 60th Annual Meeting of the Association
  for Computational Linguistics (Volume 2: Short Papers)}, pages 1--9, Dublin,
  Ireland. Association for Computational Linguistics.

\bibitem[{Cai et~al.(2020)Cai, Gan, Zhu, and Han}]{cai2020tinytl}
Han Cai, Chuang Gan, Ligeng Zhu, and Song Han. 2020.
\newblock \href
  {https://proceedings.neurips.cc/paper/2020/hash/81f7acabd411274fcf65ce2070ed568a-Abstract.html}
  {{TinyTL}: Reduce memory, not parameters for efficient on-device learning}.
\newblock In \emph{Advances in Neural Information Processing Systems},
  volume~33.

\bibitem[{Carlini et~al.(2022)Carlini, Ippolito, Jagielski, Lee, Tramer, and
  Zhang}]{carlini2022quantifying}
Nicholas Carlini, Daphne Ippolito, Matthew Jagielski, Katherine Lee, Florian
  Tramer, and Chiyuan Zhang. 2022.
\newblock \href {https://arxiv.org/abs/2202.07646} {Quantifying memorization
  across neural language models}.
\newblock \emph{arXiv preprint arXiv:2202.07646}.

\bibitem[{Carlini et~al.(2021)Carlini, Tramèr, Wallace, Jagielski,
  Herbert-Voss, Lee, Roberts, Brown, Song, Úlfar Erlingsson, Oprea, and
  Raffel}]{carlini2021extracting}
Nicholas Carlini, Florian Tramèr, Eric Wallace, Matthew Jagielski, Ariel
  Herbert-Voss, Katherine Lee, Adam Roberts, Tom~B. Brown, Dawn Song, Úlfar
  Erlingsson, Alina Oprea, and Colin Raffel. 2021.
\newblock \href
  {https://www.usenix.org/conference/usenixsecurity21/presentation/carlini-extracting}
  {Extracting training data from large language models}.
\newblock In \emph{USENIX Security Symposium}, pages 2633--2650.

\bibitem[{Devlin et~al.(2019)Devlin, Chang, Lee, and
  Toutanova}]{devlin-etal-2019-bert}
Jacob Devlin, Ming-Wei Chang, Kenton Lee, and Kristina Toutanova. 2019.
\newblock \href {https://doi.org/10.18653/v1/N19-1423} {{BERT}: Pre-training of
  deep bidirectional transformers for language understanding}.
\newblock In \emph{Proceedings of the 2019 Conference of the North {A}merican
  Chapter of the Association for Computational Linguistics: Human Language
  Technologies, Volume 1 (Long and Short Papers)}, pages 4171--4186,
  Minneapolis, Minnesota. Association for Computational Linguistics.

\bibitem[{Dwork et~al.(2014)Dwork, Roth et~al.}]{dwork2014algorithmic}
Cynthia Dwork, Aaron Roth, et~al. 2014.
\newblock \href {https://www.nowpublishers.com/article/Details/TCS-042} {The
  algorithmic foundations of differential privacy}.
\newblock \emph{Foundations and Trends{\textregistered} in Theoretical Computer
  Science}, 9(3--4):211--407.

\bibitem[{Gheini et~al.(2021)Gheini, Ren, and May}]{gheini-etal-2021-cross}
Mozhdeh Gheini, Xiang Ren, and Jonathan May. 2021.
\newblock \href {https://doi.org/10.18653/v1/2021.emnlp-main.132}
  {Cross-attention is all you need: {A}dapting pretrained {T}ransformers for
  machine translation}.
\newblock In \emph{Proceedings of the 2021 Conference on Empirical Methods in
  Natural Language Processing}, pages 1754--1765, Online and Punta Cana,
  Dominican Republic. Association for Computational Linguistics.

\bibitem[{Gopi et~al.(2021)Gopi, Lee, and Wutschitz}]{gopi2021numerical}
Sivakanth Gopi, Yin~Tat Lee, and Lukas Wutschitz. 2021.
\newblock \href
  {https://proceedings.neurips.cc/paper/2021/file/6097d8f3714205740f30debe1166744e-Paper.pdf}
  {Numerical composition of differential privacy}.
\newblock In \emph{Advances in Neural Information Processing Systems},
  volume~34, pages 11631--11642. Curran Associates, Inc.

\bibitem[{Hambardzumyan et~al.(2021)Hambardzumyan, Khachatrian, and
  May}]{hambardzumyan-etal-2021-warp}
Karen Hambardzumyan, Hrant Khachatrian, and Jonathan May. 2021.
\newblock \href {https://doi.org/10.18653/v1/2021.acl-long.381} {{WARP}:
  {W}ord-level {A}dversarial {R}e{P}rogramming}.
\newblock In \emph{Proceedings of the 59th Annual Meeting of the Association
  for Computational Linguistics and the 11th International Joint Conference on
  Natural Language Processing (Volume 1: Long Papers)}, pages 4921--4933,
  Online. Association for Computational Linguistics.

\bibitem[{He et~al.(2022)He, Zhou, Ma, Berg-Kirkpatrick, and
  Neubig}]{he2022towards}
Junxian He, Chunting Zhou, Xuezhe Ma, Taylor Berg-Kirkpatrick, and Graham
  Neubig. 2022.
\newblock \href {https://openreview.net/forum?id=0RDcd5Axok} {Towards a unified
  view of parameter-efficient transfer learning}.
\newblock In \emph{International Conference on Learning Representations}.

\bibitem[{Houlsby et~al.(2019)Houlsby, Giurgiu, Jastrzebski, Morrone,
  De~Laroussilhe, Gesmundo, Attariyan, and Gelly}]{houlsby2019parameter}
Neil Houlsby, Andrei Giurgiu, Stanislaw Jastrzebski, Bruna Morrone, Quentin
  De~Laroussilhe, Andrea Gesmundo, Mona Attariyan, and Sylvain Gelly. 2019.
\newblock \href {http://proceedings.mlr.press/v97/houlsby19a.html}
  {Parameter-efficient transfer learning for {NLP}}.
\newblock In \emph{International Conference on Machine Learning}, pages
  2790--2799. PMLR.

\bibitem[{Hu et~al.(2022)Hu, yelong shen, Wallis, Allen-Zhu, Li, Wang, Wang,
  and Chen}]{hu2022lora}
Edward~J Hu, yelong shen, Phillip Wallis, Zeyuan Allen-Zhu, Yuanzhi Li, Shean
  Wang, Lu~Wang, and Weizhu Chen. 2022.
\newblock \href {https://openreview.net/forum?id=nZeVKeeFYf9} {Lo{RA}: Low-rank
  adaptation of large language models}.
\newblock In \emph{International Conference on Learning Representations}.

\bibitem[{Karimi~Mahabadi et~al.(2021)Karimi~Mahabadi, Henderson, and
  Ruder}]{karimi2021compacter}
Rabeeh Karimi~Mahabadi, James Henderson, and Sebastian Ruder. 2021.
\newblock \href {https://openreview.net/forum?id=bqGK5PyI6-N} {Compacter:
  Efficient low-rank hypercomplex adapter layers}.
\newblock \emph{Advances in Neural Information Processing Systems},
  34:1022--1035.

\bibitem[{Lester et~al.(2021)Lester, Al-Rfou, and
  Constant}]{lester-etal-2021-power}
Brian Lester, Rami Al-Rfou, and Noah Constant. 2021.
\newblock \href {https://doi.org/10.18653/v1/2021.emnlp-main.243} {The power of
  scale for parameter-efficient prompt tuning}.
\newblock In \emph{Proceedings of the 2021 Conference on Empirical Methods in
  Natural Language Processing}, pages 3045--3059, Online and Punta Cana,
  Dominican Republic. Association for Computational Linguistics.

\bibitem[{Li et~al.(2018)Li, Farkhoor, Liu, and Yosinski}]{li2018measuring}
Chunyuan Li, Heerad Farkhoor, Rosanne Liu, and Jason Yosinski. 2018.
\newblock \href {https://openreview.net/forum?id=ryup8-WCW} {Measuring the
  intrinsic dimension of objective landscapes}.
\newblock In \emph{International Conference on Learning Representations}.

\bibitem[{Li and Liang(2021)}]{li-liang-2021-prefix}
Xiang~Lisa Li and Percy Liang. 2021.
\newblock \href {https://doi.org/10.18653/v1/2021.acl-long.353} {Prefix-tuning:
  Optimizing continuous prompts for generation}.
\newblock In \emph{Proceedings of the 59th Annual Meeting of the Association
  for Computational Linguistics and the 11th International Joint Conference on
  Natural Language Processing (Volume 1: Long Papers)}, pages 4582--4597,
  Online. Association for Computational Linguistics.

\bibitem[{Li et~al.(2022)Li, Tramer, Liang, and Hashimoto}]{li2022large}
Xuechen Li, Florian Tramer, Percy Liang, and Tatsunori Hashimoto. 2022.
\newblock \href {https://openreview.net/forum?id=bVuP3ltATMz} {Large language
  models can be strong differentially private learners}.
\newblock In \emph{International Conference on Learning Representations}.

\bibitem[{Liu et~al.(2022{\natexlab{a}})Liu, Yuan, Fu, Jiang, Hayashi, and
  Neubig}]{liu2021pre}
Pengfei Liu, Weizhe Yuan, Jinlan Fu, Zhengbao Jiang, Hiroaki Hayashi, and
  Graham Neubig. 2022{\natexlab{a}}.
\newblock \href {https://doi.org/10.1145/3560815} {Pre-train, prompt, and
  predict: A systematic survey of prompting methods in natural language
  processing}.
\newblock \emph{ACM Comput. Surv.}
\newblock Just Accepted.

\bibitem[{Liu et~al.(2022{\natexlab{b}})Liu, Ji, Fu, Tam, Du, Yang, and
  Tang}]{liu-etal-2022-p}
Xiao Liu, Kaixuan Ji, Yicheng Fu, Weng Tam, Zhengxiao Du, Zhilin Yang, and Jie
  Tang. 2022{\natexlab{b}}.
\newblock \href {https://doi.org/10.18653/v1/2022.acl-short.8} {{P}-tuning:
  Prompt tuning can be comparable to fine-tuning across scales and tasks}.
\newblock In \emph{Proceedings of the 60th Annual Meeting of the Association
  for Computational Linguistics (Volume 2: Short Papers)}, pages 61--68,
  Dublin, Ireland. Association for Computational Linguistics.

\bibitem[{Liu et~al.(2019)Liu, Ott, Goyal, Du, Joshi, Chen, Levy, Lewis,
  Zettlemoyer, and Stoyanov}]{liu2019roberta}
Yinhan Liu, Myle Ott, Naman Goyal, Jingfei Du, Mandar Joshi, Danqi Chen, Omer
  Levy, Mike Lewis, Luke Zettlemoyer, and Veselin Stoyanov. 2019.
\newblock \href {https://arxiv.org/abs/1907.11692} {{RoBERTa}: A robustly
  optimized bert pretraining approach}.
\newblock \emph{arXiv preprint arXiv:1907.11692}.

\bibitem[{Mao et~al.(2022)Mao, Mathias, Hou, Almahairi, Ma, Han, Yih, and
  Khabsa}]{mao-etal-2022-unipelt}
Yuning Mao, Lambert Mathias, Rui Hou, Amjad Almahairi, Hao Ma, Jiawei Han,
  Scott Yih, and Madian Khabsa. 2022.
\newblock \href {https://doi.org/10.18653/v1/2022.acl-long.433} {{U}ni{PELT}: A
  unified framework for parameter-efficient language model tuning}.
\newblock In \emph{Proceedings of the 60th Annual Meeting of the Association
  for Computational Linguistics (Volume 1: Long Papers)}, pages 6253--6264,
  Dublin, Ireland. Association for Computational Linguistics.

\bibitem[{Narayan et~al.(2018)Narayan, Cohen, and
  Lapata}]{narayan-etal-2018-dont}
Shashi Narayan, Shay~B. Cohen, and Mirella Lapata. 2018.
\newblock \href {https://doi.org/10.18653/v1/D18-1206} {Don{'}t give me the
  details, just the summary! topic-aware convolutional neural networks for
  extreme summarization}.
\newblock In \emph{Proceedings of the 2018 Conference on Empirical Methods in
  Natural Language Processing}, pages 1797--1807, Brussels, Belgium.
  Association for Computational Linguistics.

\bibitem[{Pfeiffer et~al.(2021)Pfeiffer, Kamath, R{\"u}ckl{\'e}, Cho, and
  Gurevych}]{pfeiffer-etal-2021-adapterfusion}
Jonas Pfeiffer, Aishwarya Kamath, Andreas R{\"u}ckl{\'e}, Kyunghyun Cho, and
  Iryna Gurevych. 2021.
\newblock \href {https://doi.org/10.18653/v1/2021.eacl-main.39}
  {{A}dapter{F}usion: Non-destructive task composition for transfer learning}.
\newblock In \emph{Proceedings of the 16th Conference of the European Chapter
  of the Association for Computational Linguistics: Main Volume}, pages
  487--503, Online. Association for Computational Linguistics.

\bibitem[{Pfeiffer et~al.(2020)Pfeiffer, R{\"u}ckl{\'e}, Poth, Kamath,
  Vuli{\'c}, Ruder, Cho, and Gurevych}]{pfeiffer-etal-2020-adapterhub}
Jonas Pfeiffer, Andreas R{\"u}ckl{\'e}, Clifton Poth, Aishwarya Kamath, Ivan
  Vuli{\'c}, Sebastian Ruder, Kyunghyun Cho, and Iryna Gurevych. 2020.
\newblock \href {https://doi.org/10.18653/v1/2020.emnlp-demos.7}
  {{A}dapter{H}ub: A framework for adapting transformers}.
\newblock In \emph{Proceedings of the 2020 Conference on Empirical Methods in
  Natural Language Processing: System Demonstrations}, pages 46--54, Online.
  Association for Computational Linguistics.

\bibitem[{Radford et~al.(2019)Radford, Wu, Child, Luan, Amodei, Sutskever
  et~al.}]{radford2019language}
Alec Radford, Jeffrey Wu, Rewon Child, David Luan, Dario Amodei, Ilya
  Sutskever, et~al. 2019.
\newblock Language models are unsupervised multitask learners.
\newblock \emph{OpenAI blog}, 1(8):9.

\bibitem[{Ro et~al.(2022)Ro, Breiner, McConnaughey, Chen, Suresh, Kumar, and
  Mathews}]{ro2022scaling}
Jae~Hun Ro, Theresa Breiner, Lara McConnaughey, Mingqing Chen, Ananda~Theertha
  Suresh, Shankar Kumar, and Rajiv Mathews. 2022.
\newblock \href {https://openreview.net/forum?id=ShNG29KGF-c} {Scaling language
  model size in cross-device federated learning}.
\newblock In \emph{ACL 2022 Workshop on Federated Learning for Natural Language
  Processing}.

\bibitem[{R{\"u}ckl{\'e} et~al.(2021)R{\"u}ckl{\'e}, Geigle, Glockner, Beck,
  Pfeiffer, Reimers, and Gurevych}]{ruckle-etal-2021-adapterdrop}
Andreas R{\"u}ckl{\'e}, Gregor Geigle, Max Glockner, Tilman Beck, Jonas
  Pfeiffer, Nils Reimers, and Iryna Gurevych. 2021.
\newblock \href {https://doi.org/10.18653/v1/2021.emnlp-main.626}
  {{AdapterDrop}: {O}n the efficiency of adapters in transformers}.
\newblock In \emph{Proceedings of the 2021 Conference on Empirical Methods in
  Natural Language Processing}, pages 7930--7946, Online and Punta Cana,
  Dominican Republic. Association for Computational Linguistics.

\bibitem[{Schick and Sch{\"u}tze(2021)}]{schick-schutze-2021-just}
Timo Schick and Hinrich Sch{\"u}tze. 2021.
\newblock \href {https://doi.org/10.18653/v1/2021.naacl-main.185} {It{'}s not
  just size that matters: Small language models are also few-shot learners}.
\newblock In \emph{Proceedings of the 2021 Conference of the North American
  Chapter of the Association for Computational Linguistics: Human Language
  Technologies}, pages 2339--2352, Online. Association for Computational
  Linguistics.

\bibitem[{Shin et~al.(2020)Shin, Razeghi, Logan~IV, Wallace, and
  Singh}]{shin-etal-2020-autoprompt}
Taylor Shin, Yasaman Razeghi, Robert~L. Logan~IV, Eric Wallace, and Sameer
  Singh. 2020.
\newblock \href {https://doi.org/10.18653/v1/2020.emnlp-main.346}
  {{A}uto{P}rompt: {E}liciting {K}nowledge from {L}anguage {M}odels with
  {A}utomatically {G}enerated {P}rompts}.
\newblock In \emph{Proceedings of the 2020 Conference on Empirical Methods in
  Natural Language Processing (EMNLP)}, pages 4222--4235, Online. Association
  for Computational Linguistics.

\bibitem[{Subramani and Suresh(2020)}]{subramani2020discovering}
Nishant Subramani and Nivedita Suresh. 2020.
\newblock \href {https://arxiv.org/abs/2008.09049} {Discovering useful sentence
  representations from large pretrained language models}.
\newblock \emph{arXiv preprint arXiv:2008.09049}.

\bibitem[{Tjong Kim~Sang and
  De~Meulder(2003)}]{tjong-kim-sang-de-meulder-2003-introduction}
Erik~F. Tjong Kim~Sang and Fien De~Meulder. 2003.
\newblock \href {https://aclanthology.org/W03-0419} {Introduction to the
  {C}o{NLL}-2003 shared task: Language-independent named entity recognition}.
\newblock In \emph{Proceedings of the Seventh Conference on Natural Language
  Learning at {HLT}-{NAACL} 2003}, pages 142--147.

\bibitem[{Tramer and Boneh(2021)}]{tramer2021differentially}
Florian Tramer and Dan Boneh. 2021.
\newblock \href {https://openreview.net/forum?id=YTWGvpFOQD-} {Differentially
  private learning needs better features (or much more data)}.
\newblock In \emph{International Conference on Learning Representations}.

\bibitem[{Treviso et~al.(2022)Treviso, Ji, Lee, van Aken, Cao, Ciosici, Hassid,
  Heafield, Hooker, Martins, Martins, Milder, Raffel, Simpson, Slonim,
  Balasubramanian, Derczynski, and Schwartz}]{treviso2022efficient}
Marcos Treviso, Tianchu Ji, Ji-Ung Lee, Betty van Aken, Qingqing Cao, Manuel~R.
  Ciosici, Michael Hassid, Kenneth Heafield, Sara Hooker, Pedro~H. Martins,
  André F.~T. Martins, Peter Milder, Colin Raffel, Edwin Simpson, Noam Slonim,
  Niranjan Balasubramanian, Leon Derczynski, and Roy Schwartz. 2022.
\newblock \href {https://arxiv.org/abs/2209.00099} {Efficient methods for
  natural language processing: A survey}.
\newblock \emph{arXiv preprint arXiv:2209.00099}.

\bibitem[{Wang et~al.(2019)Wang, Singh, Michael, Hill, Levy, and
  Bowman}]{wang2018glue}
Alex Wang, Amanpreet Singh, Julian Michael, Felix Hill, Omer Levy, and
  Samuel~R. Bowman. 2019.
\newblock \href {https://openreview.net/forum?id=rJ4km2R5t7} {{GLUE}: A
  multi-task benchmark and analysis platform for natural language
  understanding}.
\newblock In \emph{International Conference on Learning Representations}.

\bibitem[{Wolf et~al.(2020)Wolf, Debut, Sanh, Chaumond, Delangue, Moi, Cistac,
  Rault, Louf, Funtowicz, Davison, Shleifer, von Platen, Ma, Jernite, Plu, Xu,
  Le~Scao, Gugger, Drame, Lhoest, and Rush}]{wolf-etal-2020-transformers}
Thomas Wolf, Lysandre Debut, Victor Sanh, Julien Chaumond, Clement Delangue,
  Anthony Moi, Pierric Cistac, Tim Rault, Remi Louf, Morgan Funtowicz, Joe
  Davison, Sam Shleifer, Patrick von Platen, Clara Ma, Yacine Jernite, Julien
  Plu, Canwen Xu, Teven Le~Scao, Sylvain Gugger, Mariama Drame, Quentin Lhoest,
  and Alexander Rush. 2020.
\newblock \href {https://doi.org/10.18653/v1/2020.emnlp-demos.6} {Transformers:
  State-of-the-art natural language processing}.
\newblock In \emph{Proceedings of the 2020 Conference on Empirical Methods in
  Natural Language Processing: System Demonstrations}, pages 38--45, Online.
  Association for Computational Linguistics.

\bibitem[{Xu et~al.(2022)Xu, Song, Tian, Agrawal, Granqvist, van Dalen, Zhang,
  Argueta, Han, Deng et~al.}]{xu2022training}
Mingbin Xu, Congzheng Song, Ye~Tian, Neha Agrawal, Filip Granqvist, Rogier van
  Dalen, Xiao Zhang, Arturo Argueta, Shiyi Han, Yaqiao Deng, et~al. 2022.
\newblock \href {https://arxiv.org/abs/2207.08988} {Training large-vocabulary
  neural language models by private federated learning for resource-constrained
  devices}.
\newblock \emph{arXiv preprint arXiv:2207.08988}.

\bibitem[{Yang et~al.(2022)Yang, Lin, Yang, Wang, Zhou, and
  Yang}]{yang2022prompt}
Hao Yang, Junyang Lin, An~Yang, Peng Wang, Chang Zhou, and Hongxia Yang. 2022.
\newblock \href {https://arxiv.org/abs/2208.02532} {Prompt tuning for
  generative multimodal pretrained models}.
\newblock \emph{arXiv preprint arXiv:2208.02532}.

\bibitem[{Yousefpour et~al.(2021)Yousefpour, Shilov, Sablayrolles, Testuggine,
  Prasad, Malek, Nguyen, Ghosh, Bharadwaj, Zhao, Cormode, and Mironov}]{opacus}
Ashkan Yousefpour, Igor Shilov, Alexandre Sablayrolles, Davide Testuggine,
  Karthik Prasad, Mani Malek, John Nguyen, Sayan Ghosh, Akash Bharadwaj,
  Jessica Zhao, Graham Cormode, and Ilya Mironov. 2021.
\newblock \href {https://arxiv.org/abs/2109.12298} {Opacus: {U}ser-friendly
  differential privacy library in {PyTorch}}.
\newblock \emph{arXiv preprint arXiv:2109.12298}.

\bibitem[{Yu et~al.(2022)Yu, Naik, Backurs, Gopi, Inan, Kamath, Kulkarni, Lee,
  Manoel, Wutschitz, Yekhanin, and Zhang}]{yu2022differentially}
Da~Yu, Saurabh Naik, Arturs Backurs, Sivakanth Gopi, Huseyin~A Inan, Gautam
  Kamath, Janardhan Kulkarni, Yin~Tat Lee, Andre Manoel, Lukas Wutschitz,
  Sergey Yekhanin, and Huishuai Zhang. 2022.
\newblock \href {https://openreview.net/forum?id=Q42f0dfjECO} {Differentially
  private fine-tuning of language models}.
\newblock In \emph{International Conference on Learning Representations}.

\bibitem[{Zhou et~al.(2022{\natexlab{a}})Zhou, Xu, and
  McAuley}]{zhou2022efficiently}
Wangchunshu Zhou, Canwen Xu, and Julian McAuley. 2022{\natexlab{a}}.
\newblock \href {https://arxiv.org/abs/2210.11705} {Efficiently tuned
  parameters are task embeddings}.
\newblock \emph{arXiv preprint arXiv:2210.11705}.

\bibitem[{Zhou et~al.(2022{\natexlab{b}})Zhou, Ma, Zou, Chen, Gui, Zhang,
  Huang, Xie, and Wu}]{zhou-etal-2022-making}
Xin Zhou, Ruotian Ma, Yicheng Zou, Xuanting Chen, Tao Gui, Qi~Zhang, Xuanjing
  Huang, Rui Xie, and Wei Wu. 2022{\natexlab{b}}.
\newblock \href {https://aclanthology.org/2022.coling-1.615} {Making
  parameter-efficient tuning more efficient: A unified framework for
  classification tasks}.
\newblock In \emph{Proceedings of the 29th International Conference on
  Computational Linguistics}, pages 7053--7064, Gyeongju, Republic of Korea.
  International Committee on Computational Linguistics.

\end{thebibliography}
